\documentclass{article}

     \PassOptionsToPackage{numbers, compress}{natbib}
     \usepackage[preprint]{neurips_2022}

\usepackage[utf8]{inputenc} % allow utf-8 input
\usepackage[T1]{fontenc}    % use 8-bit T1 fonts
\usepackage{hyperref}       % hyperlinks
\usepackage{url}            % simple URL typesetting
\usepackage{booktabs}       % professional-quality tables
\usepackage{amsfonts}       % blackboard math symbols
\usepackage{nicefrac}       % compact symbols for 1/2, etc.
\usepackage{microtype}      % microtypography
\usepackage{xcolor}         % colors

\usepackage{makecell}
\usepackage{multirow}
\usepackage{multicol}

\usepackage{epsfig}
\usepackage{amsmath}
\usepackage{amssymb}
\usepackage{wrapfig}

\usepackage{comment}
\usepackage[accsupp]{axessibility}

\usepackage{float}
\usepackage{pifont}

\newcommand{\xmark}{\ding{55}}
\usepackage{caption}
\usepackage{subcaption}

\title{Contrastive Learning Rivals Masked Image Modeling in Fine-tuning via Feature Distillation}
\author{%
  Yixuan Wei\thanks{Equal. $^\dag$Correspondence. Yixuan and Zhenda are long-term interns at MSRA.}\textsuperscript{~~12} \quad Han Hu\textsuperscript{*$\dag$2} \quad Zhenda Xie\textsuperscript{12} \quad Zheng Zhang\textsuperscript{2} \NewAnd Yue Cao\textsuperscript{2} \quad Jianmin Bao\textsuperscript{2} \quad Dong Chen\textsuperscript{2} \quad Baining Guo\textsuperscript{2} \\
  $^1$Tsinghua University \quad 
  $^2$Microsoft Research Asia\\
  \texttt{\{t-yixuanwei,hanhu,t-zhxie,zhez,yuecao,jianbao,doch,bainguo\}@microsoft.com} \\
}

\begin{document}

\maketitle

\begin{abstract}

Masked image modeling (MIM) learns representations with remarkably good fine-tuning performances, overshadowing previous prevalent pre-training approaches such as image classification, instance contrastive learning, and image-text alignment. In this paper, we show that the inferior fine-tuning performance of these pre-training approaches can be significantly improved by a simple post-processing in the form of feature distillation (FD). The feature distillation converts the old representations to new representations that have a few desirable properties just like those representations produced by MIM. These properties, which we aggregately refer to as optimization friendliness, are identified and analyzed by a set of attention- and optimization-related diagnosis tools. With these properties, the new representations show strong fine-tuning performance. Specifically, the contrastive self-supervised learning methods are made as competitive in fine-tuning as the state-of-the-art masked image modeling (MIM) algorithms. The CLIP models' fine-tuning performance is also significantly improved, with a CLIP ViT-L model reaching \textbf{89.0\%} top-1 accuracy on ImageNet-1K classification. On the 3-billion-parameter SwinV2-G model, the fine-tuning accuracy is improved by +1.5 mIoU / +1.1 mAP to \textbf{61.4 mIoU} / \textbf{64.2 mAP} on ADE20K semantic segmentation and COCO object detection, respectively, creating new records on both benchmarks. More importantly, our work provides a way for the future research to focus more effort on the generality and scalability of the learnt representations without being pre-occupied with optimization friendliness since it can be enhanced rather easily. The code will be available at \url{https://github.com/SwinTransformer/Feature-Distillation}.

\end{abstract}

\section{Introduction}

The paradigm of pre-training and fine-tuning has played a key role in the development of deep learning methods in the field of computer vision. In 2006, the autoencoder based pre-training~\cite{HinSal06} is a pioneer work that largely triggered the burst of deep learning. Furthermore, since AlexNet~\cite{alexnet} achieved revolutionary recognition accuracy on ImageNet-1K image classification~\cite{deng2009imagenet} in 2012, model pre-training using the image classification task has become a standard practice for a variety of down-stream computer vision tasks, including object detection~\cite{rcnn13} and semantic segmentation~\cite{long2015fully}.

For representation learning in computer vision, two notable approaches have been quite successful: the instance contrastive learning~\cite{dosovitskiy2014exemplarcnn,he2019moco,chen2020simclr,grill2020byol,chen2021mocov3,caron2021emerging} and the image-text alignment methods~\cite{radford2021clip,jia2021align}. The former learns representation in a self-supervised manner, and achieves impressive linear evaluation performance on image classification~\cite{he2019moco,caron2021emerging}. The latter, represented by the CLIP approach~\cite{radford2021clip}, is notable for opening up the field of zero-shot recognition, allowing visual recognition models to classify almost any category. However, when fine-tuned on down-stream vision tasks, their performance is generally not superior to other methods~\cite{xie2021self, caron2021emerging,li2021esvit,ftclip2021} and thus limits their wider adoption. 

\begin{table}[t]
\caption{Feature distillation improves fine-tuning performance. * uses the same model in the upper row, but with an additional inter-mediate fine-tuning step on ImageNet-22K image classification.}
\centering
  \begin{tabular}{l|c|c|c|cc|c}
\Xhline{1.0pt}
  \multirow{2}{*}{Method} & \multirow{2}{*}{Backbone} & \multirow{2}{*}{res.} & \multirow{2}{*}{F. D.} &  \multicolumn{2}{c|}{IN-1K} & \multirow{2}{*}{ADE20K}  \\
  \cline{5-6}
   &  &  &  & f.t. & linear   &   \\
  \hline
  \color{gray}{BEiT~\cite{bao2021beit}} & \color{gray}{ViT-B} & \color{gray}{$224^2$} & & \color{gray}{83.2} & \color{gray}{37.6} & \color{gray}{47.1} \\
  \color{gray}{MAE~\cite{MaskedAutoencoders2021}} & \color{gray}{ViT-B} & \color{gray}{$224^2$} & & \color{gray}{83.6} & \color{gray}{68.0} & \color{gray}{48.1} \\
  \color{gray}{SimMIM~\cite{xie2021simmim}} & \color{gray}{ViT-B} & \color{gray}{$224^2$} & & \color{gray}{83.8} & \color{gray}{56.7} & \color{gray}{47.6} \\
  \color{gray}{SimMIM~\cite{xie2021simmim}} & \color{gray}{Swin-B} & \color{gray}{$224^2$} & & \color{gray}{84.8} & \color{gray}{24.8} & \color{gray}{48.3} \\
   \hline
\color{gray}{WiSE-FT CLIP~\cite{ftclip2021}} & \color{gray}{ViT-L} & \color{gray}{336$^2$} & \color{gray}{} & \color{gray}{87.1} & \color{gray}{-} & \color{gray}{-}\\
  \hline
  \hline
  DINO~\cite{caron2021emerging} & ViT-B & $224^2$ & & 82.8 & 78.2 & 46.2 \\
  FD-DINO & ViT-B & $224^2$ & $\checkmark$ & \textbf{83.8}\scriptsize{ (+1.0)} & 76.1 &  \textbf{47.7}\scriptsize{ (+1.5)}\\
  \hline
  \hline
  EsViT~\cite{li2021esvit} & Swin-B & $224^2$ & & 83.9 & 81.3 & 47.3 \\
  FD-EsViT & Swin-B & $224^2$ & $\checkmark$ & \textbf{85.1}\scriptsize{ (+1.2)} & 80.4 & \textbf{48.9}\scriptsize{ (+1.6)}\\
  \hline
  \hline
  DeiT~\cite{touvron2020deit} & \multirow{2}{*}{ViT-B} & \multirow{1}{*}{$224^2$} & & 81.8 & - & 47.0 \\
  FD-DeiT &  & \multirow{1}{*}{$224^2$} & $\checkmark$ & \textbf{83.0}\scriptsize{ (+1.2)} & - & \textbf{48.0} \scriptsize{ (+1.0)} \\
  \hline
    \hline
  CLIP~\cite{radford2021clip} & \multirow{2}{*}{ViT-B} & \multirow{1}{*}{$224^2$} &  & 82.9 & 79.5 & 49.5 \\
  FD-CLIP & & \multirow{1}{*}{$224^2$} & $\checkmark$ & \textbf{84.9}\scriptsize{ (+2.0)} & 80.3 & \textbf{52.8}\scriptsize{ (+3.3)} \\
  \hline
  CLIP~\cite{radford2021clip} & \multirow{3}{*}{ViT-L} & \multirow{1}{*}{$224^2$} & & 86.1 & 83.5 & 53.5\\
  FD-CLIP &  & $224^2$ & $\checkmark$ & \textbf{87.7}\scriptsize{ (+1.6)} & 84.8 & \textbf{55.7}\scriptsize{ (+2.2)}\\
  FD-CLIP* & & 336$^2$ & $\checkmark$ & \textbf{89.0}\scriptsize{} &  - & - \\
\Xhline{1.0pt}
  \end{tabular}
\label{tab:teaser_FD}
\end{table}

\begin{table}[t]
\caption{Feature distillation improves the state-of-the-art SwinV2-G model.}
\centering
  \begin{tabular}{l|c|c|c|c}
\Xhline{1.0pt}
  \multirow{1}{*}{Method} &  \multirow{1}{*}{IN-1K (\%)} & \multirow{1}{*}{COCO (AP$_\text{box}$)} & \multirow{1}{*}{COCO (AP$_\text{mask}$)} & \multirow{1}{*}{ADE20K (mIoU)} \\
  \hline
  \color{gray}{GLIPv2-CoSwin-H}~\cite{GLIPv2_2022} & \color{gray}{-} & \color{gray}{62.4} & \color{gray}{-} & \color{gray}{-} \\
  \color{gray}{Florence-CoSwin-H}~\cite{yuan2021florence} & \color{gray}{-} & \color{gray}{62.4} & \color{gray}{-} & \color{gray}{-}  \\
  \color{gray}{DINO-Swin-L}~\cite{zhang2022dino} & \color{gray}{-} & \color{gray}{63.3} & \color{gray}{-} & \color{gray}{-} \\
  \color{gray}{MaskDINO-Swin-L}~\cite{li2022mask} & \color{gray}{-} & \color{gray}{-} & \color{gray}{54.7} & \color{gray}{60.8} \\
  \color{gray}{ViT-Adapter-L}~\cite{chen2022vitadapter} & \color{gray}{-} & \color{gray}{-} & \color{gray}{-} & \color{gray}{60.5} \\
    \hline
  \hline
  SwinV2-G~\cite{swinv2} & 89.2 & 63.1 & 54.4 & 59.9 \\
  FD-SwinV2-G & \textbf{89.4}\scriptsize{ (+0.2)} & \textbf{64.2}\scriptsize{ (+1.1)} & \textbf{55.4}\scriptsize{ (+1.0)} & \textbf{61.4}\scriptsize{ (+1.5)} \\
\Xhline{1.0pt}
  \end{tabular}
\label{tab:FD_swinv2_G}
\end{table}
Recently, the masked image modeling (MIM)~\cite{chen2020imagegpt,bao2021beit,xie2021simmim,MaskedAutoencoders2021} has achieved remarkable performance in fine-tuning evaluations~\cite{bao2021beit,xie2021simmim,MaskedAutoencoders2021} and attracted widespread attention. The success of MIM begs for the question: why does MIM perform so much better in fine-tuning? In other words, are there key ingredients that can be added to other pre-training approaches to make them as successful as MIM in fine-tuning? 

In this paper, we show that a simple feature distillation method can generally improve the fine-tuning performance of various pre-training methods, including contrastive based self-supervised learning approaches such as DINO~\cite{caron2021emerging} and EsViT~\cite{li2021esvit}, visual-language models such as CLIP~\cite{radford2021clip}, and image classification methods such as DeiT~\cite{touvron2021training}, as shown in Table~\ref{tab:teaser_FD}. In our feature distillation method, the already learnt representations are distilled into new features that are trained from scratch. For the distillation targets, we advocate feature maps rather than logits, which allows the FD approach to handle features obtained by arbitrary pre-training methods and leads to better fine-tuning accuracy. We also propose useful designs that are beneficial for the successive fine-tuning process, including whitened distillation targets, shared relative position bias, and asymmetric drop path rates. With this approach and the careful design, contrastive based self-supervised pre-training approaches such as DINO and EsViT become competitive in the fine-tuning evaluation as the masked image modeling approaches, or even slightly better. The CLIP pre-trained ViT-L model achieves an 89.0\% top-1 accuracy on ImageNet-1K image classification, a new state-of-the-art result with ViT-L. On the 3-billion-parameter SwinV2-G model, the fine-tuning accuracy is improved by +1.5 mIoU / +1.1 mAP to \textbf{61.4 mIoU} / \textbf{64.2 mAP} on ADE20K semantic segmentation and COCO object detection, respectively, creating new records on both benchmarks (see Table~\ref{tab:FD_swinv2_G}).

We analyze the model properties before and after feature distillation with a set of attention- and optimization-related diagnostic tools. We observe that
the feature distillation converts the old representations to new ones embodied a few desirable properties just like those representations produced by MIM. These properties, which we aggregately refer to as optimization friendliness, are identified and analyzed by the attention- and optimization-related diagnosis tools, including average attention distances~\cite{dosovitskiy2020vit}, average attention maps, attention similarities between heads~\cite{zhou2021deepvit}, and normalized loss landscape~\cite{losslandscape2017}. These tools show that the models after distillation have more diverse attention heads, and account more on relative positions than the absolute positions. These are desirable properties that encourage flatter loss landscapes in fine-tuning as well as better final accuracy. Note that representations learnt by the masked image modeling methods already have nice optimization friendliness properties and hence adding an additional feature distillation post-processing yield little gains. Such results suggest that optimization friendly representations are the reasons behind the superior fine-tuning performance of masked image modeling methods.

Our work provides a way for the future research to focus more attention on the generality and scalability of the learnt representations without being pre-occupied with optimization friendliness. The generality and scalability are critical because they not only make the pre-training suitable for broad visual tasks but also allow the trained network to take full advantage of larger model capacity and larger data. In existing research, the goal of generality and scalability is often intertwined with that of optimization friendliness. The feature distillation approach sheds lights on how we might be able to decouple these two goals and allows more effort devoted to the critical issue of generality and scalability.

\begin{figure}
    \centering
    \includegraphics[width=\linewidth]{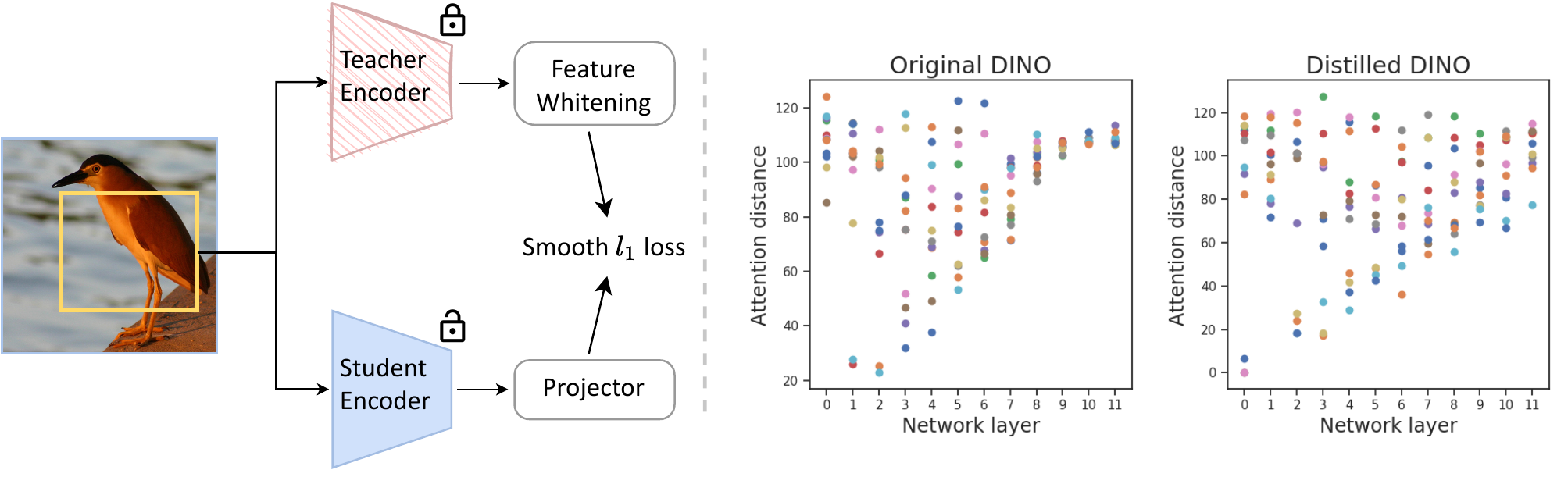}
    \caption{\textbf{Left}: Illustration of the feature distillation approach. \textbf{Right}: Average attention distances per layer and head.}
    \label{fig:teaser}
\end{figure}

\section{A Feature Distillation Method}

\label{sec:FD}

With an already pre-trained model, our goal is to obtain a new representation that distills knowledge from the already pre-trained model while being more friendly to fine-tuning. We achieve this by a feature distillation method, as illustrated in Figure~\ref{fig:teaser} (left). In this method, the already pre-trained model plays the teacher, and the new model plays the student. We consider the following designs to make this method both generic and effective.

\paragraph{Distilling feature maps so as to be generic} Instead of distilling logits as most previous distillation works have done~\cite{hinton2015knowledge}, we adopt the output feature map of the pre-trained model as the distillation target. Using the feature map as the distilling target allows us to work with any pre-trained model that may not have a logit output. In addition to being more general, distilling the feature map also shows higher fine-tuning accuracy than using logits or the reduced single feature vector (see Table~\ref{tab:ablation_targets}).

To make feature maps of the teacher and student comparable, we adopt the same augmentation view for each original image. We also apply a $1\times 1$ convolution layer on top of the student network to allow different dimensions of output feature maps between the teacher and student, such that the method can be further generalized.

\paragraph{Whitening teacher features for distillation} Different pre-trained models may have very different orders of feature magnitudes, which will make difficulties in hyper-parameter tuning for different pre-training approaches. To solve this problem, we normalize the output feature map of the teacher network by a whitening operation, which is implemented by a non-parametric layer normalization operator without scaling and bias.

In distillation, we employ a smooth $\ell_1$ loss between the student and teacher feature maps:
\begin{equation}\label{eq:supervise}
    \mathcal{L}_\text{distill} (\mathbf{s}, \mathbf{t}) = \begin{cases}
\frac{1}{2} (g(\mathbf{s}) - \text{whiten}(\mathbf{t}))^2/\beta, & | g(\mathbf{s}) - \text{whiten}(\mathbf{t}) | \leq \beta \\
(|g(\mathbf{s})-\text{whiten}(\mathbf{t})|-\frac{1}{2}\beta), & \text{otherwise}
\end{cases},
\end{equation}
where $\beta$ is set 2.0 by default; $\mathbf{s}$ and $\mathbf{t}$ are output feature vectors of the student and teacher networks, respectively; $g$ is a $1\times 1$ convolution layer.

\paragraph{Shared relative position bias} In the original ViT~\cite{dosovitskiy2020vit}, relative position bias (RPB) did not show any benefit over the absolute position encoding (APE), and so absolute position encoding (APE) is usually used for ViT architectures~\cite{dosovitskiy2020vit,caron2021emerging,chen2021mocov3,MaskedAutoencoders2021}.

In the feature distillation framework, we re-examine the effects of position encoding configuration in the student architecture, including absolute position encoding (APE) and relative position bias (RPB)~\cite{liu2021swin}. We also consider a \emph{shared RPB} configuration, where all layers share the same relative positional bias matrices. We find that the \emph{shared RPB} performed best overall, as shown in Table~\ref{tab:ablation_position}. We find that the \emph{shared RPB} can diversify the attention distances of heads, especially for the deeper layers (see figure ~\ref{fig:att_similarity} and \ref{fig:position_configuration}), which may cause its slightly better fine-tuning accuracy. We use the \emph{shared RPB} by default in our experiments. 

\paragraph{Asymmetric drop path rates} The two-branch manner in the feature distillation framework allows us to use asymmetric regularization for the teacher and student networks. We find that a strategy of asymmetric drop path~\cite{huang2016deep} rates is beneficial for learning better representations. Specifically, on ViT-B, the strategy of applying a drop path rate of 0.1-0.3 on the student branch, and no drop path regularization on the teacher branch works best, as shown in Table~\ref{tab:ablation_dpr}.

\section{Representations before and after Feature Distillation}

In this section, we delve into the feature distillation mechanism introduced in the previous section through a set of attention- and optimization-related diagnostic tools, including the average attention distance per head~\cite{dosovitskiy2020vit}, average cosine similarities between attention maps of heads~\citep{zhou2021deepvit}, average attention maps for each layer, and normalized loss landscapes~\cite{losslandscape2017}. We perform these analyses using 50,000 ImageNet-1K validation images and diagnose the models both before and after applying the distillation method. Different property behaviors of learnt representations are observed before and after feature distillation.

\begin{figure}
     \centering
     \begin{minipage}{.50\textwidth}
      \centering
       \includegraphics[width=\linewidth]{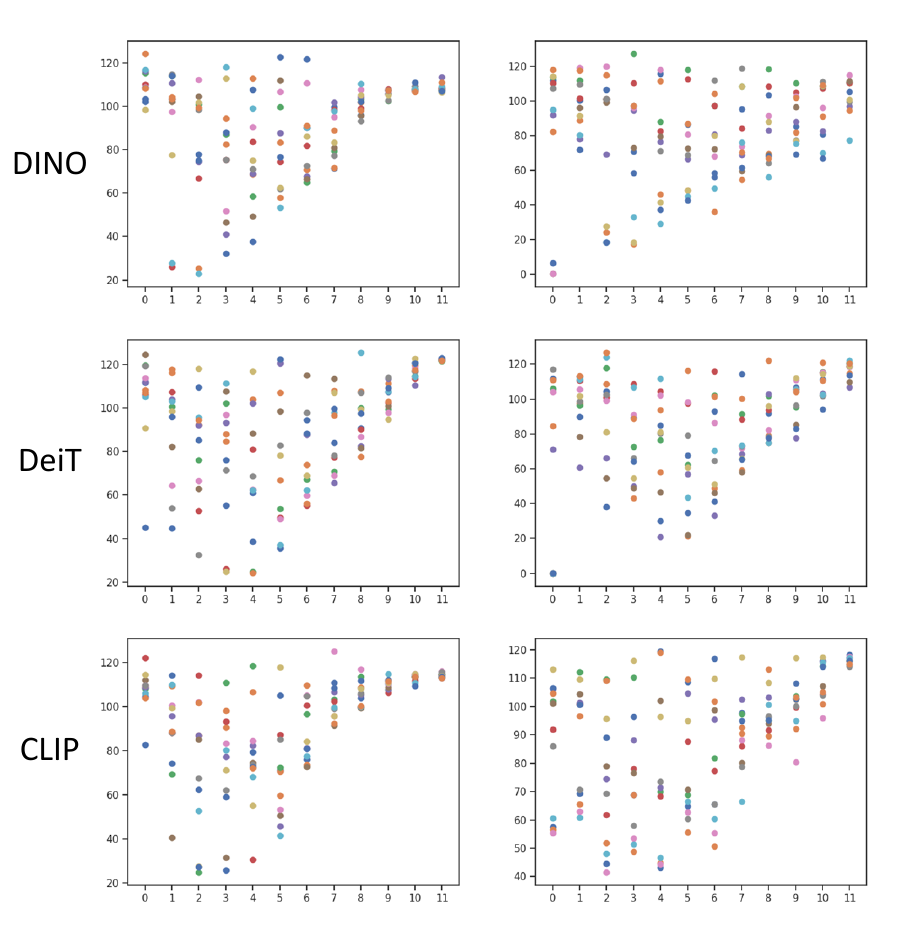}
       \caption{Average attention distances per head at each layer depth before (left) and after (right) feature distillation using ViT-B.}
       \label{fig:att_distance}
     \end{minipage}%
     \hspace{5pt}
     \begin{minipage}{.465\textwidth}
       \centering
      \includegraphics[width=\linewidth]{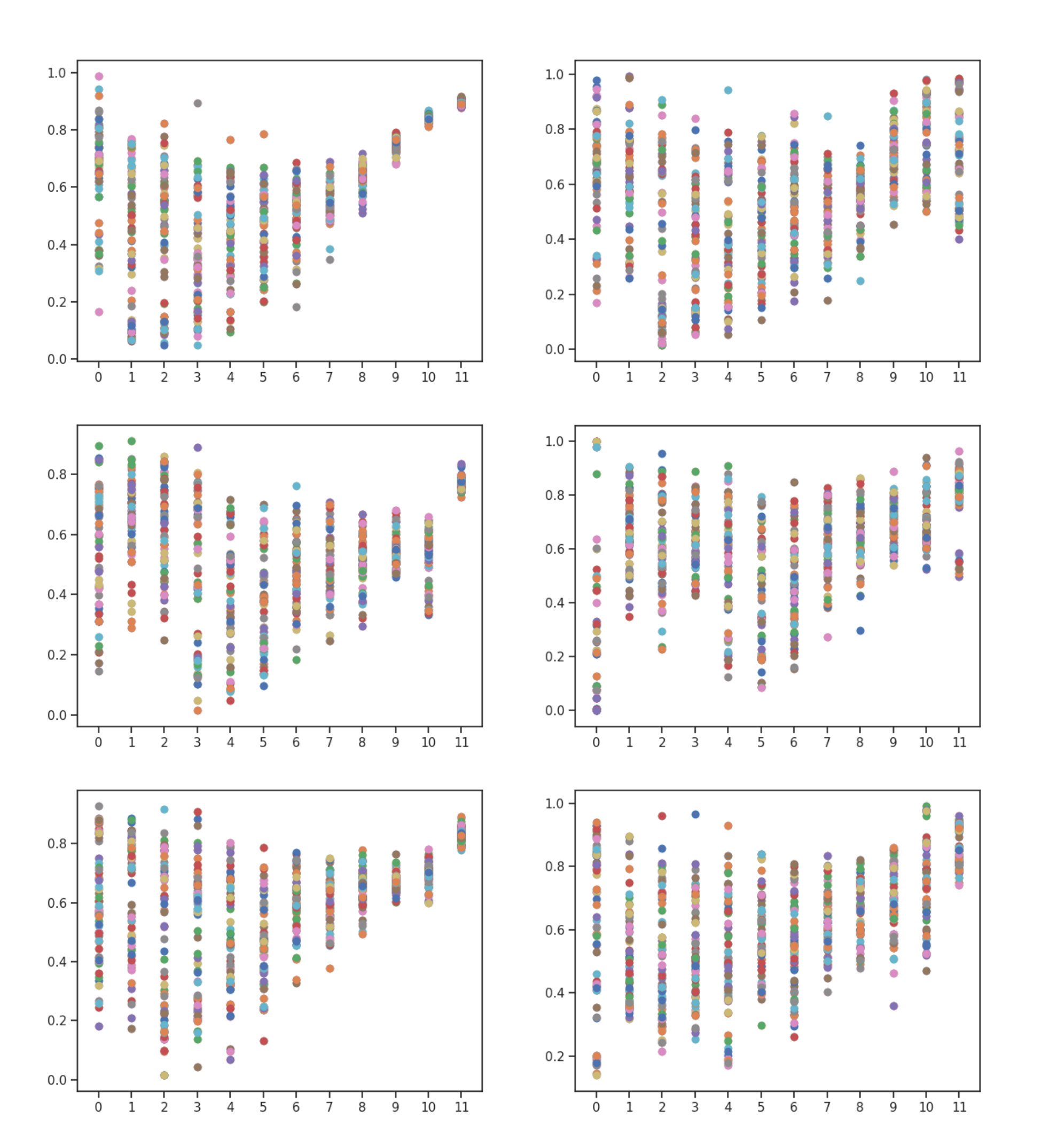}
       \caption{Average cosine similarity of attention maps between heads per layer before (left) and after (right) feature distillation using ViT-B.}
       \label{fig:att_similarity}
     \end{minipage}
 \end{figure}

\paragraph{Feature distillation diversifies attention heads} We examine the attention diversity of heads. Figure~\ref{fig:att_distance} shows the average attention distance per head and layer depth using the ViT-B architectures pretrained by DINO, DeiT, and CLIP, respectively. The average attention distance is introduced in~\cite{dosovitskiy2020vit}, which can partially reflect the receptive field size for each attention head, computed according to the attention weights. It can be seen that: For all pre-trained representations before distillation, the attention distances of different heads in deeper layers collapse to locate within a very small distance range. This suggests that different heads learn very similar visual cues and may be wasting model capacity. After the feature distillation process, all representations become more diverse or more evenly distributed regarding the attention distance, especially for deeper layers. This observation is also reflected by Figure~\ref{fig:att_similarity}, which calculates the average cosine similarity between attention heads of each layer.

\paragraph{Changes on attention patterns}

\begin{figure}
\centering
\begin{subfigure}{.5\textwidth}
  \centering
  \includegraphics[width=0.95\linewidth]{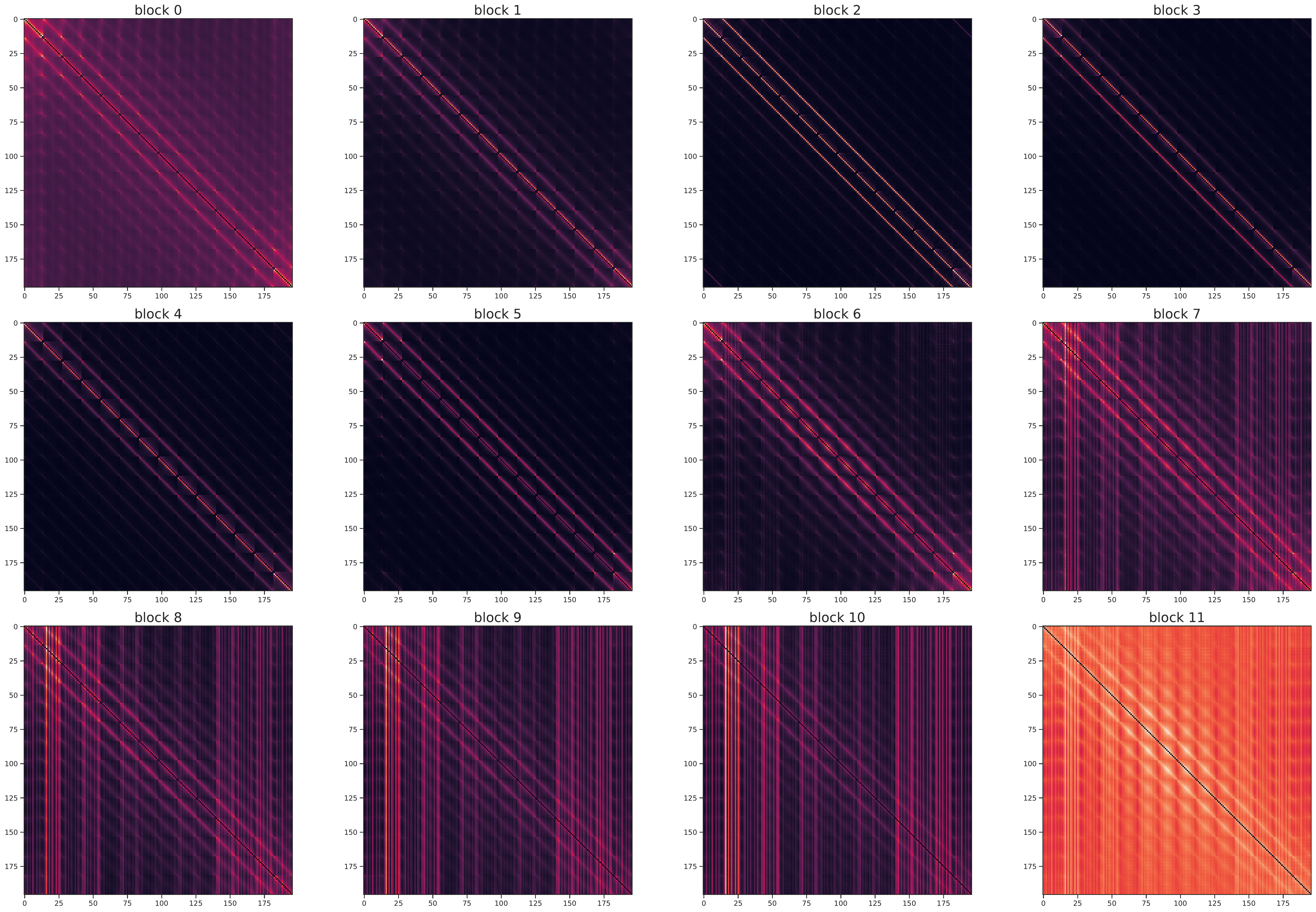}
  \caption{Before feature distillation}
  \label{fig:att_patterns_a}
\end{subfigure}%
\begin{subfigure}{.5\textwidth}
  \centering
  \includegraphics[width=0.95\linewidth]{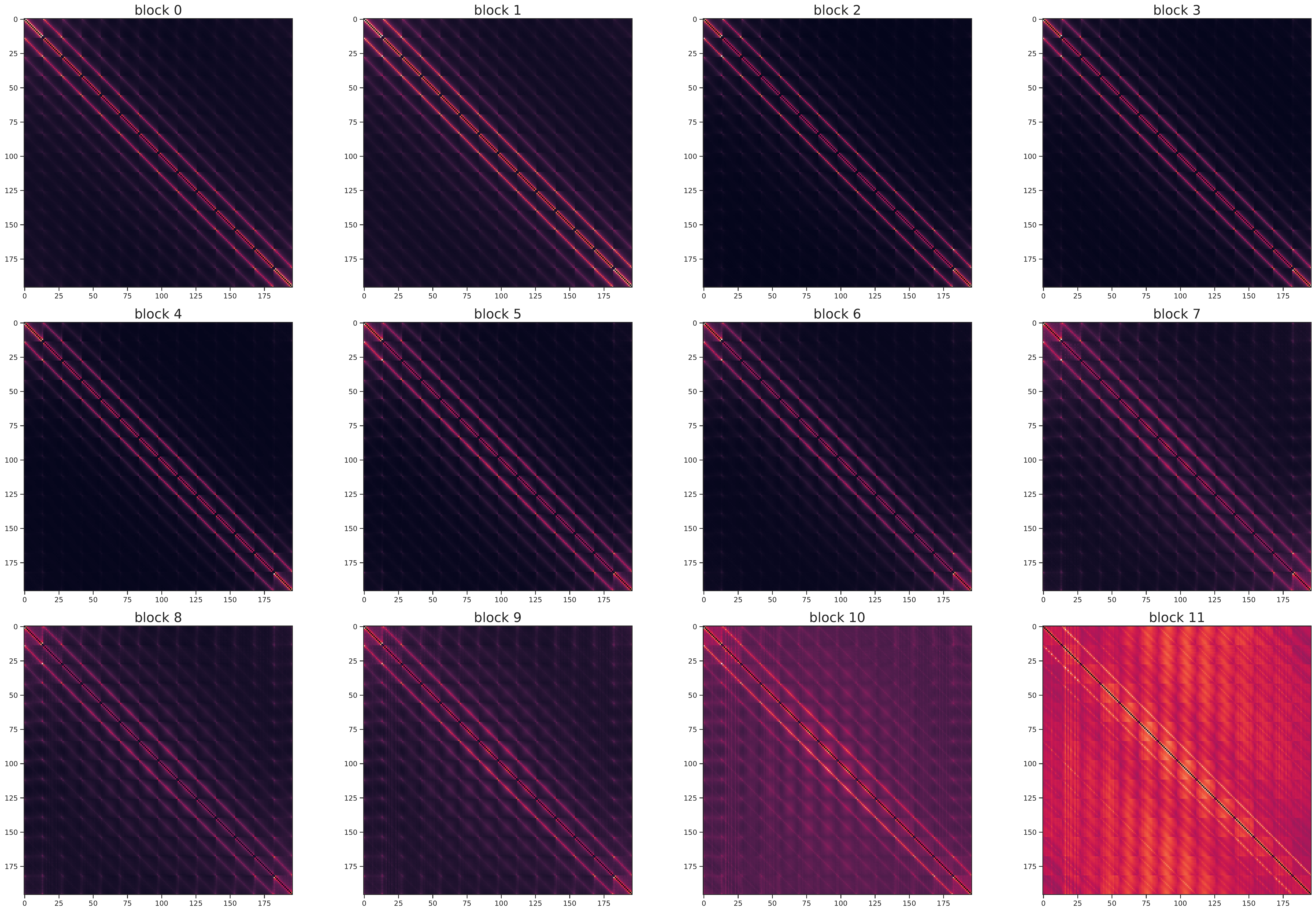}
  \caption{After feature distillation}
  \label{fig:att_patterns_b}
\end{subfigure}
    \caption{Average attention maps per layer using the CLIP ViT-B model before and after feature distillation. The image patches are indexed starting from top-left to bottom-right. The 12 layers' average attention maps (Layer 0-11) are visualized from top-left to bottom-right. The average attention maps for other pre-training approaches will be shown in appendix.}
    \label{fig:att_patterns}
\end{figure}

Figure~\ref{fig:att_patterns} shows the average attention maps before (left) and after (right) feature distillation. There are two obvious patterns in the attention maps: \emph{diagonal} and \emph{column}. The \emph{diagonal} pattern corresponds to the relationship between image patches in some fixed \emph{relative positions}, while the \emph{column} pattern represents the effect of image patches in certain absolution positions to all other locations.

It can be seen that representations after feature distillation have much more \emph{diagonal} patterns, which means the model relies more on visual cues that encode relationship of relative locations. It suggests better translational invariance of the model, which is often a beneficial property for various visual tasks. 
Noting the student network has included shared relative position bias (RPB), in order to study its effects, we also tried to use the absolute position encoding (APE) in the student architecture, whose attention maps are shown in Figure~\ref{fig:position_configuration}. In this configuration, the representations after feature distillation also rely more on relative locations, for example, Layer 0 and Layer 7, and the fine-tuning accuracy is also quite high (see Table~\ref{tab:ablation_position}). This suggests that the more \emph{diagonal} patterns are primarily caused by the feature distillation algorithm itself.

\begin{figure}
\centering
\begin{subfigure}{.24\textwidth}
  \centering
  \includegraphics[width=0.97\linewidth]{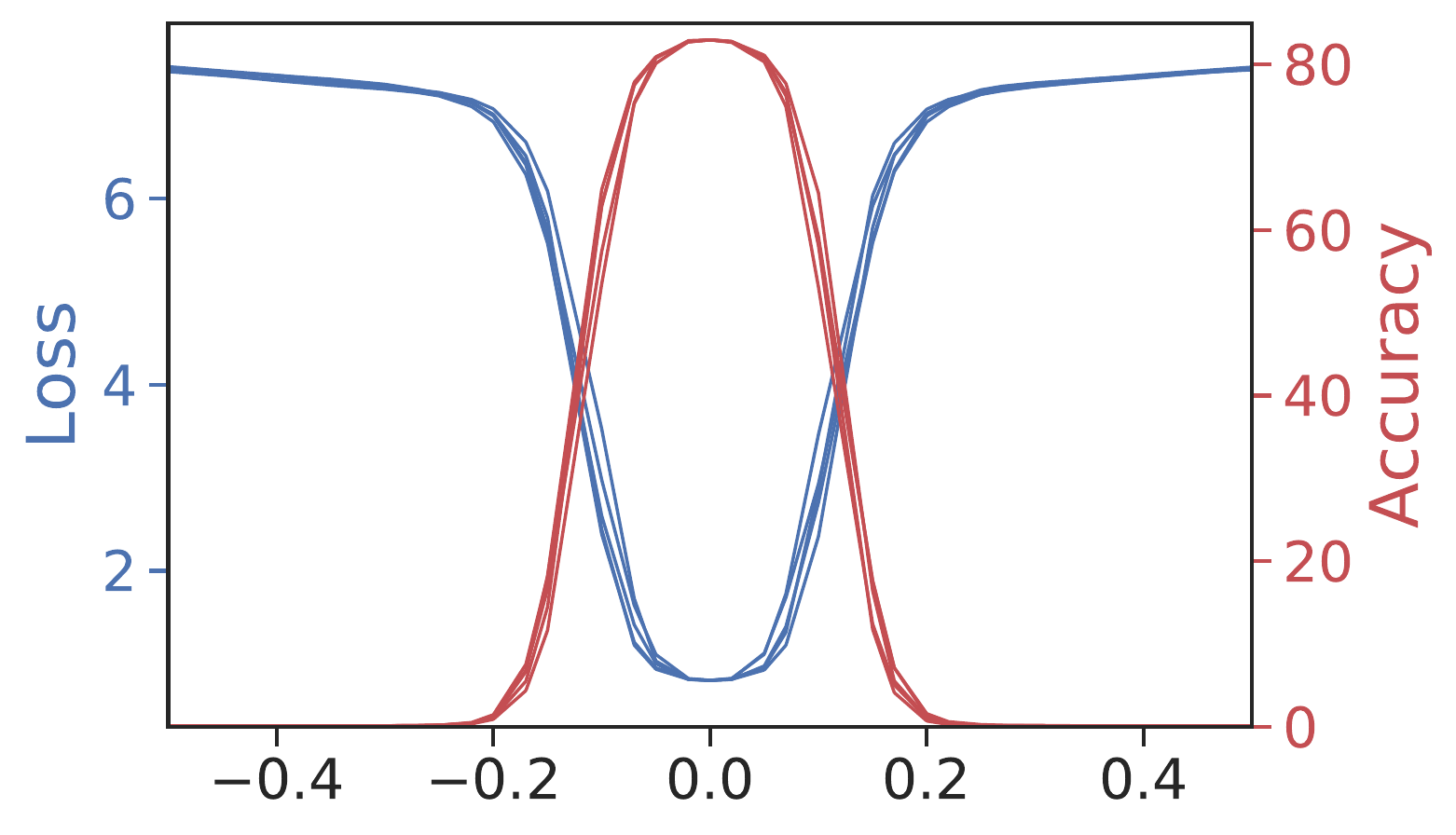}
  \label{fig:loss_landscapes_clip_a}
\end{subfigure}%
\begin{subfigure}{.24\textwidth}
  \centering
  \includegraphics[width=0.97\linewidth]{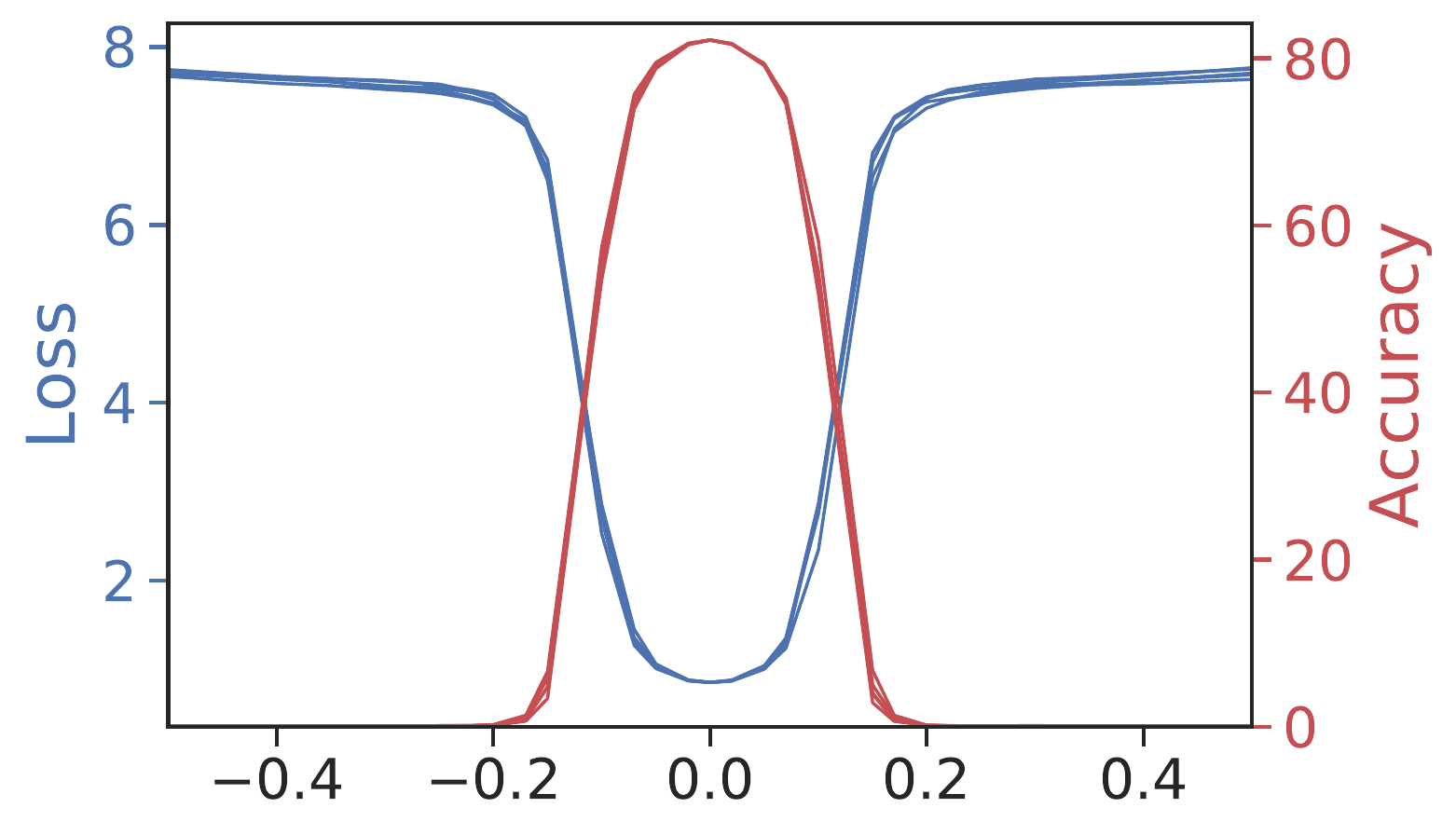}
  \label{fig:loss_landscapes_dino_a}
\end{subfigure}
\begin{subfigure}{.24\textwidth}
  \centering
  \includegraphics[width=0.97\linewidth]{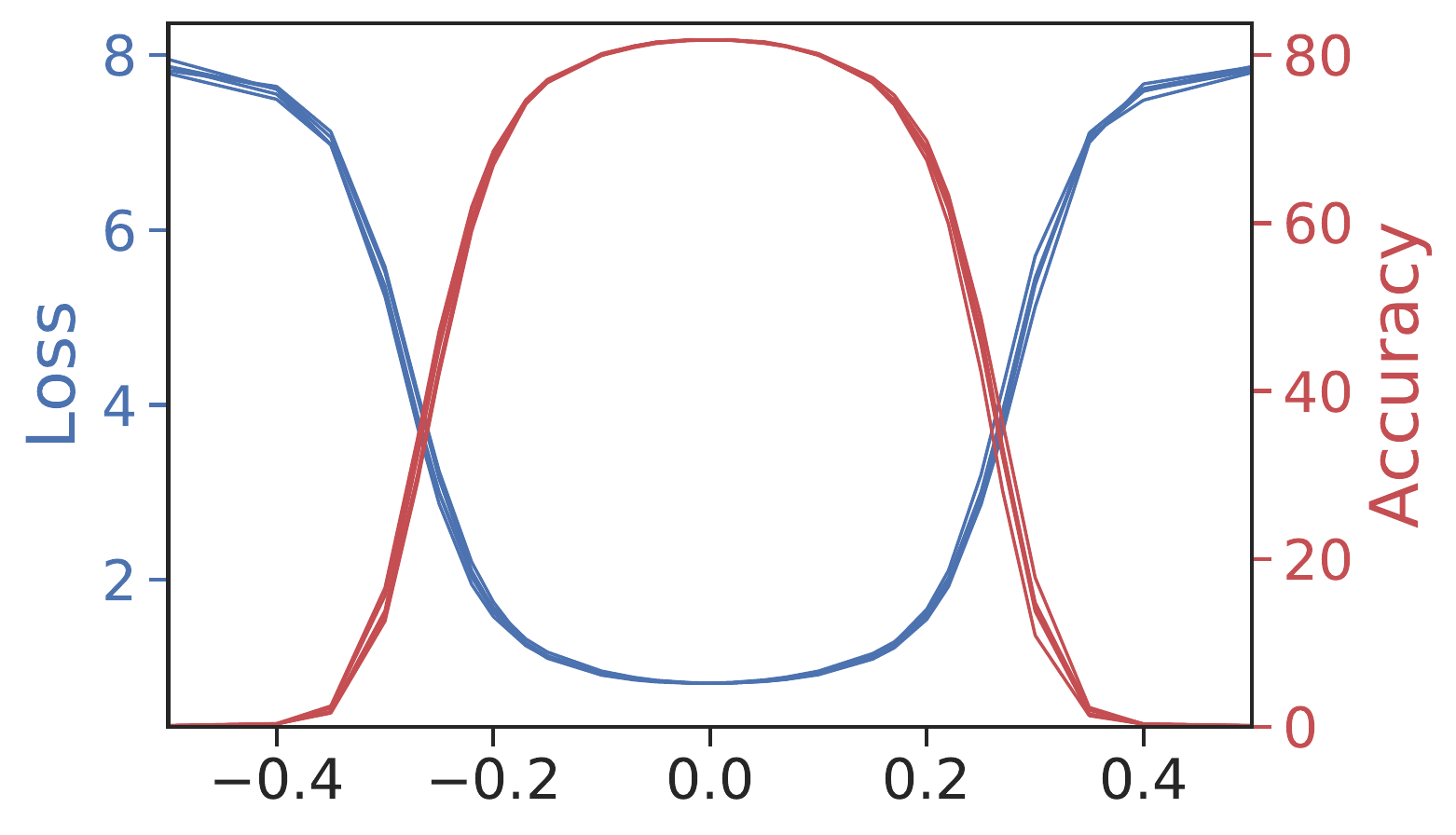}
  \label{fig:loss_landscapes_deit_a}
\end{subfigure}
\begin{subfigure}{.24\textwidth}
  \centering
  \includegraphics[width=0.97\linewidth]{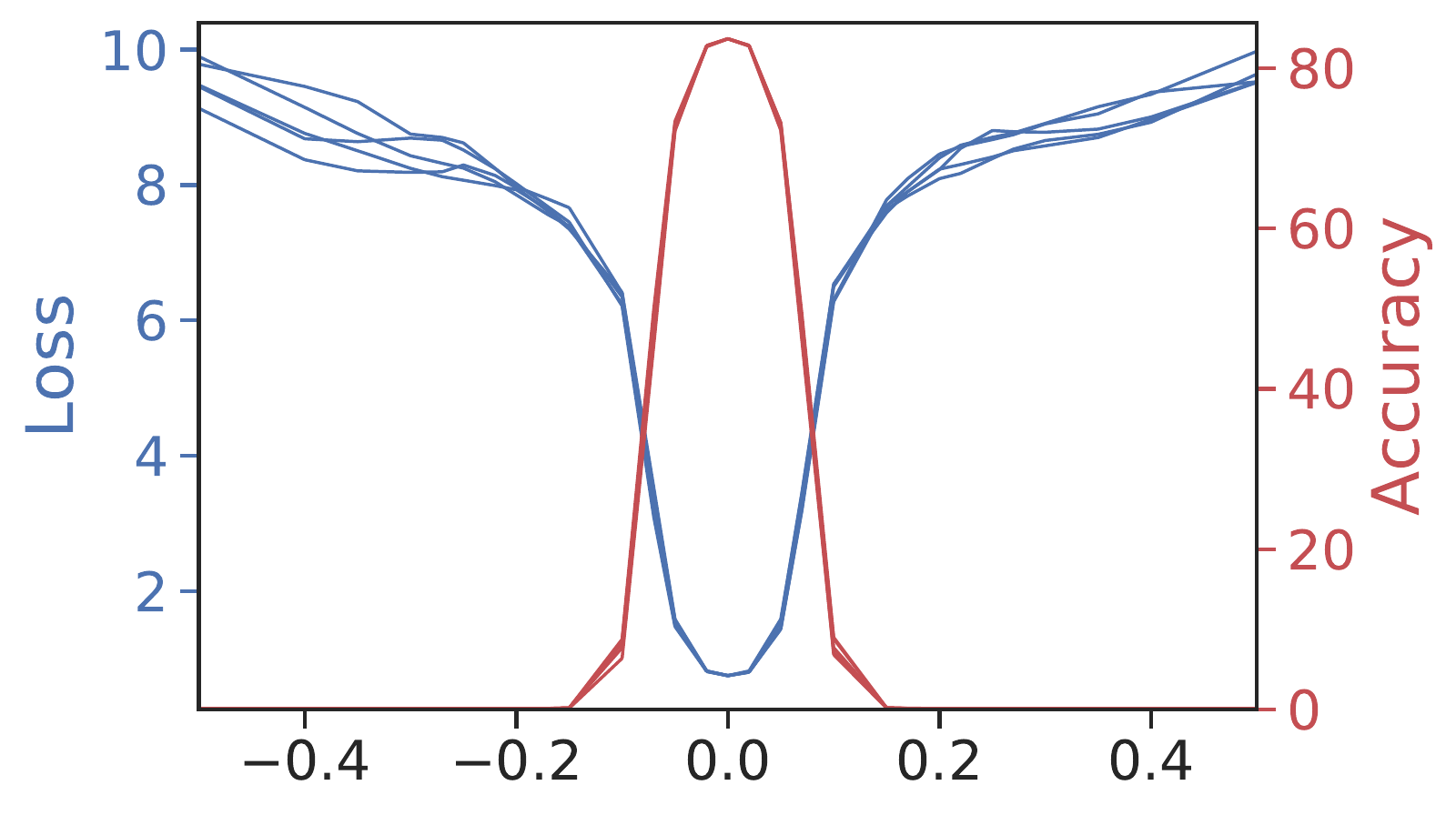}
  \label{fig:loss_landscapes_esvit_a}
\end{subfigure}

\bigskip

\begin{subfigure}{.24\textwidth}
  \centering
  \includegraphics[width=0.97\linewidth]{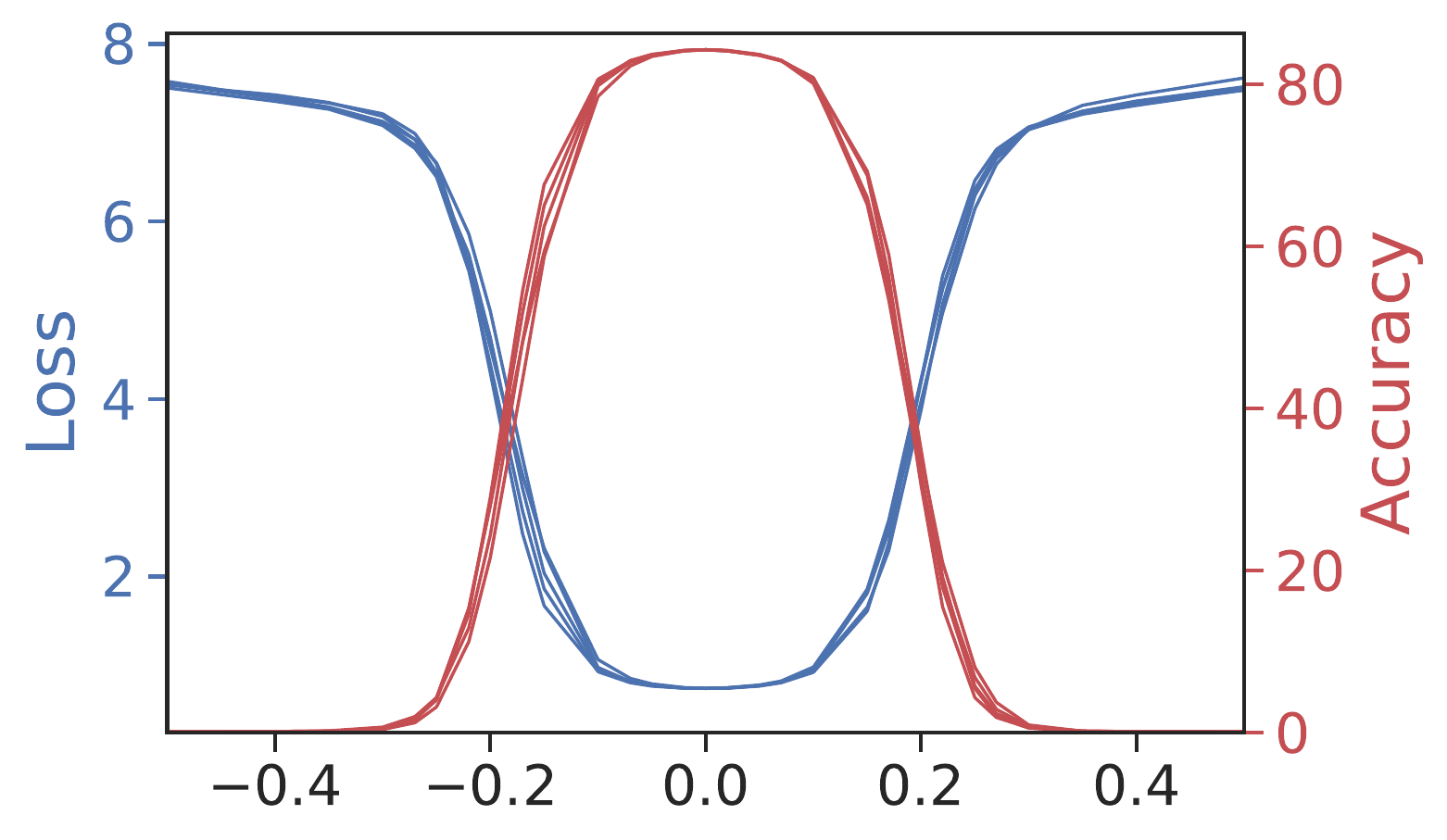}
  \caption{CLIP}
  \label{fig:loss_landscapes_clip_b}
\end{subfigure}%
\begin{subfigure}{.24\textwidth}
  \centering
  \includegraphics[width=0.97\linewidth]{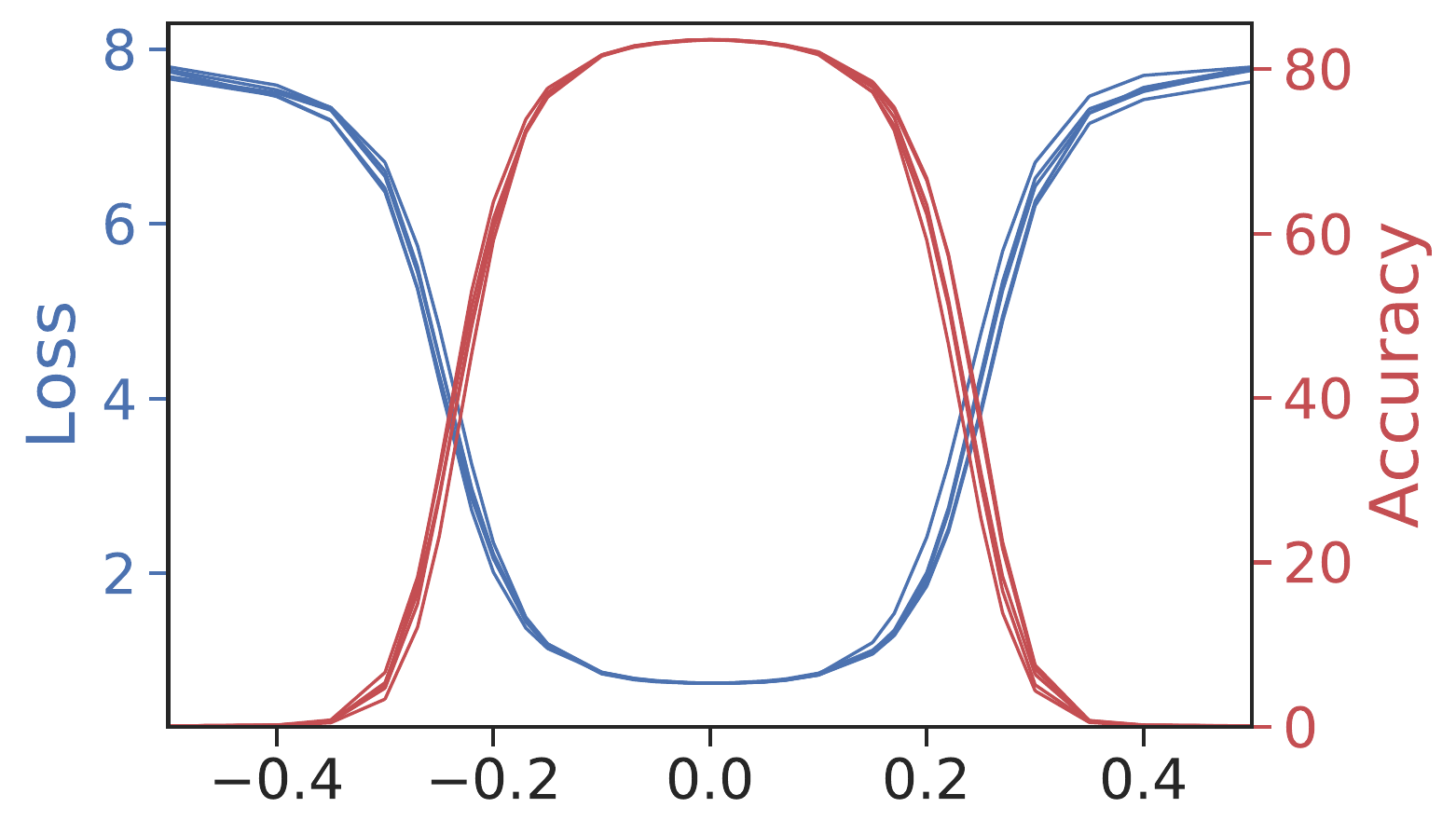}
  \caption{DINO}
  \label{fig:loss_landscapes_dino_b}
\end{subfigure}
\begin{subfigure}{.24\textwidth}
  \centering
  \includegraphics[width=0.97\linewidth]{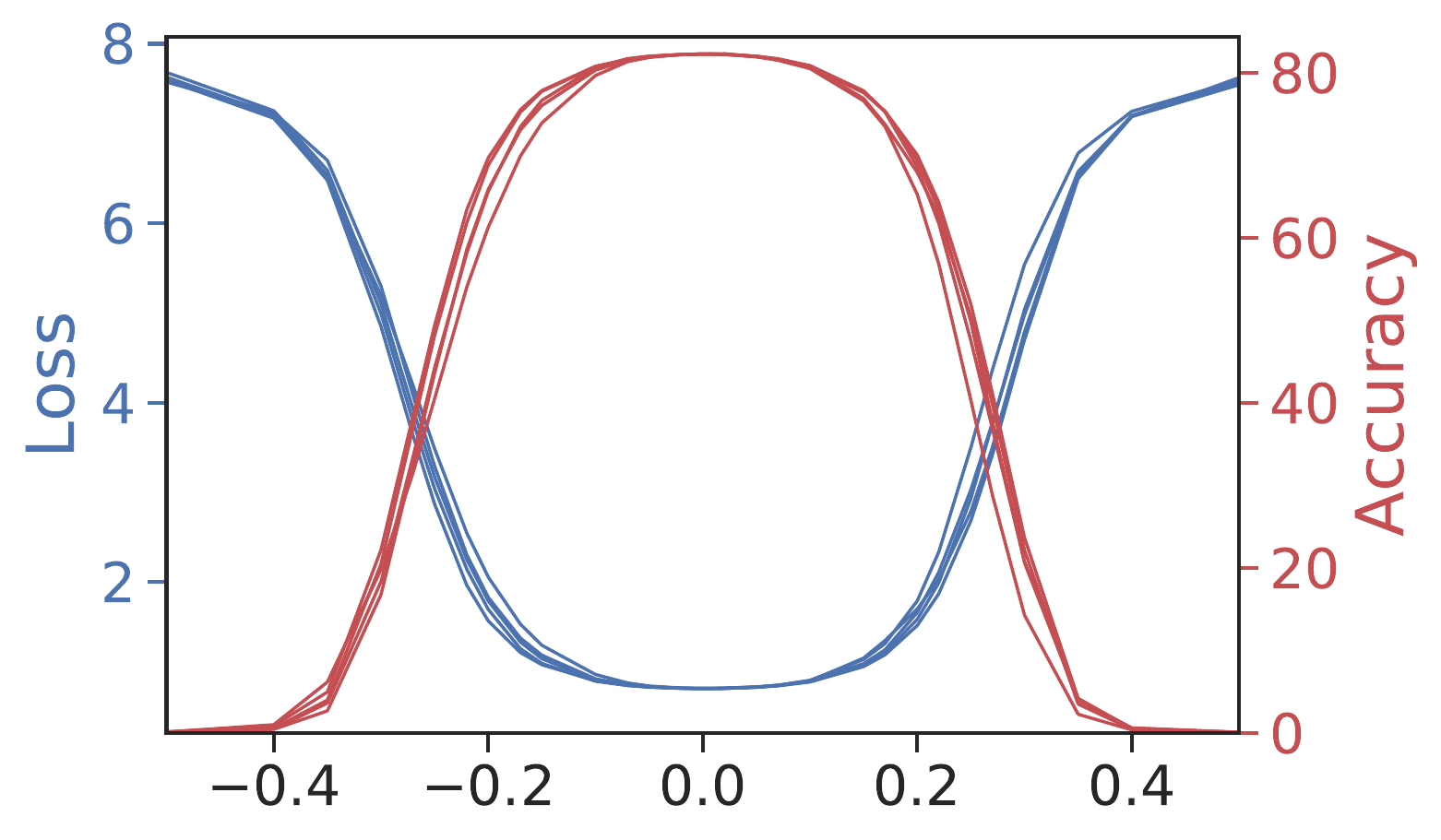}
  \caption{DeiT}
  \label{fig:loss_landscapes_deit_b}
\end{subfigure}
\begin{subfigure}{.24\textwidth}
  \centering
  \includegraphics[width=0.97\linewidth]{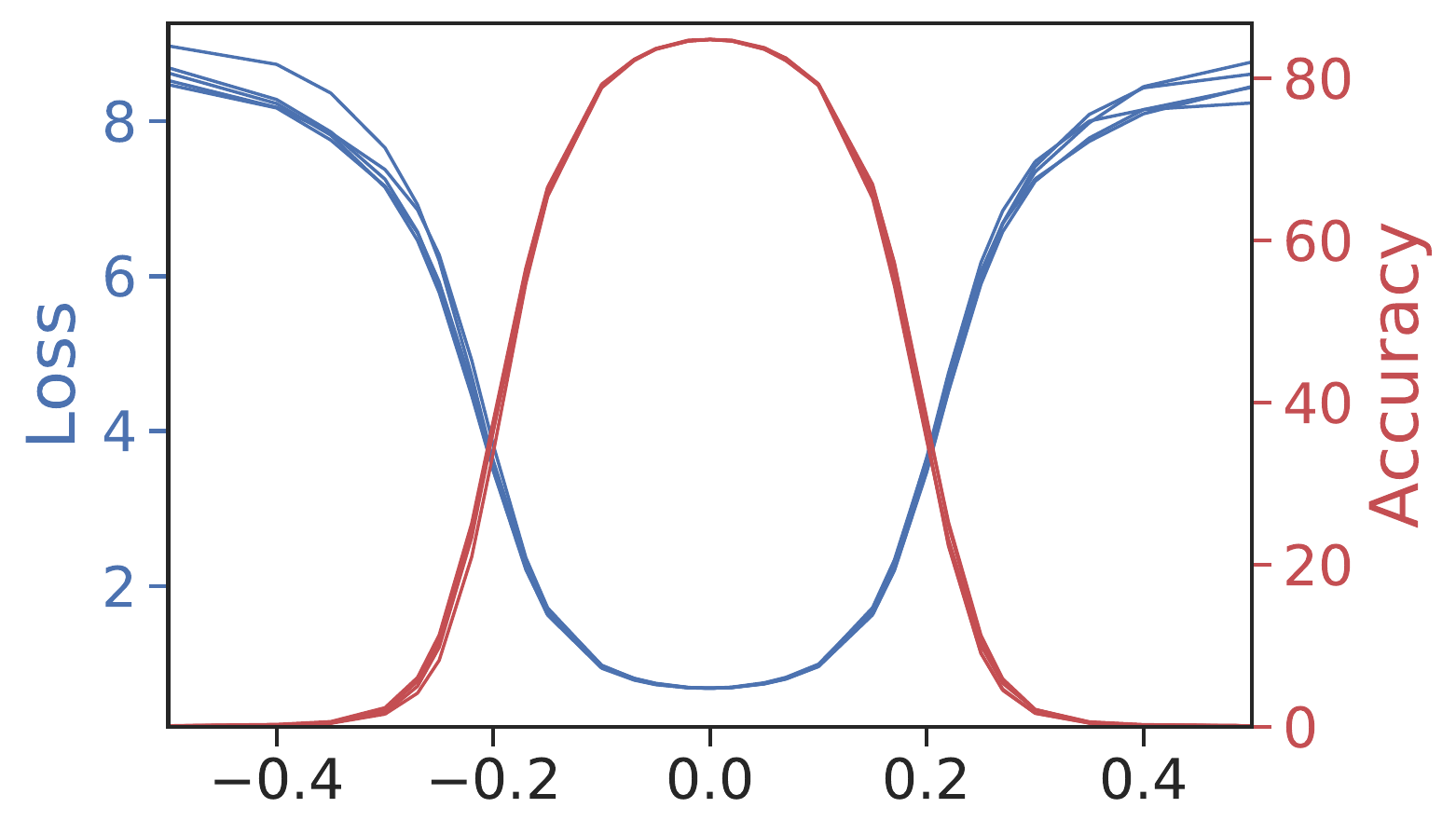}
  \caption{EsViT}
  \label{fig:loss_landscapes_esvit_b}
\end{subfigure}

    \caption{The loss / accuracy landscapes~\cite{losslandscape2017} of different pre-trained models before (top) and after (bottom) feature distillation. Each plot has 5 landscapes using 5 randomly generated directions.}
    \label{fig:loss_landscapes}
\end{figure}

\paragraph{Feature distillation gets better loss / accuracy landscapes} We use the method in~\cite{losslandscape2017} to visualize the loss / accuracy landscapes of different models. In this visualization method, the model weights are perturbed by a series of Gaussian noises with varying degrees. Following~\cite{losslandscape2017}, each noise level is defined normalized to the $\ell_2$ norm of each filter to account for the effects of varying weight amplitudes of different models. Figure~\ref{fig:loss_landscapes} visualizes the loss / accuracy landscapes of different pre-trained models before and after feature distillation. It turns out that the loss / accuracy landscapes of most representations after feature distillation have become flatter than the ones by the representations before distillation, which is consistent with their better fine-tuning accuracy. 

\begin{figure}
\centering
\begin{subfigure}{.45\textwidth}
  \centering
  \includegraphics[width=0.97\linewidth]{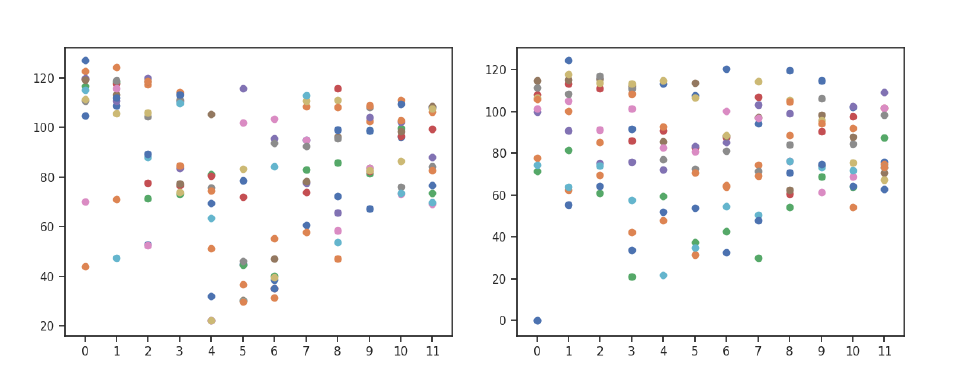}
  \caption{Average attention distance}
  \label{fig:fd_mae_a}
\end{subfigure}%
\begin{subfigure}{.55\textwidth}
  \centering
  \includegraphics[width=0.97\linewidth]{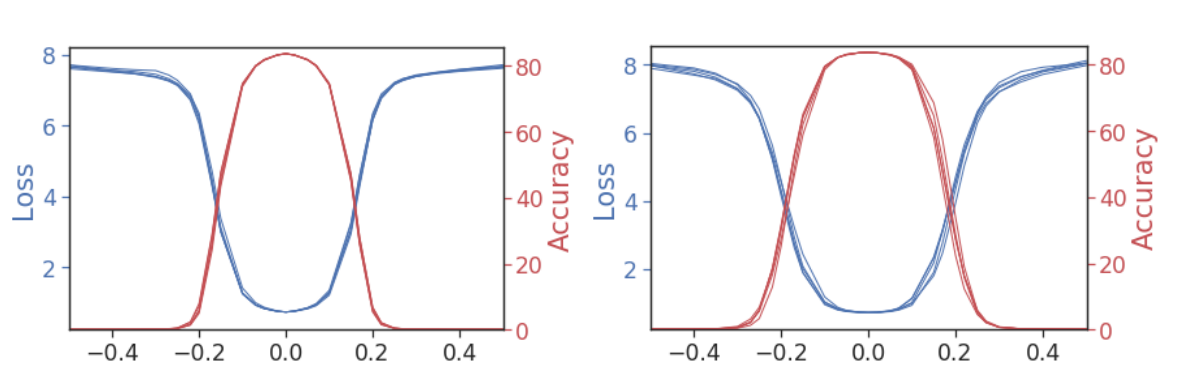}
  \caption{Accuracy and loss landscapes}
  \label{fig:fd_mae_b}
\end{subfigure}

    \caption{The average attention distances and loss / accuracy landscapes before (left) and after (right) feature distillation on a masked image modeling approach, MAE~\cite{MaskedAutoencoders2021}. }
    \label{fig:fd_mae}
\end{figure}

\paragraph{On masked image modeling (MIM)} Figure~\ref{fig:fd_mae} shows the average attention distance and the loss / accuracy landscapes of an MIM-based approach, MAE~\cite{MaskedAutoencoders2021}, before and after the feature distillation process. It can be seen that the representations pre-trained using MAE have learnt diverse heads, and that the loss / accuracy landscapes are relatively flat. In fact, an additional conversion of the old MAE representations to new ones by the feature distillation method only brings slight gains of +0.2\% (83.8\% versus 83.6\%). These results may suggest that the good fine-tuning performance brought by the feature distillation post-processing have certain overlap in functionality with the masked image modeling (MIM) method.

\section{Experiments}
\label{sec-experiment}
\subsection{Experimental Settings}

\paragraph{Distillation implementation and dataset} For all experiments, we perform feature distillation using 1.28M ImageNet-1K training images. In ablation, we distill 100 epochs for all experiments. In system-level comparison such as Table~\ref{tab:teaser_FD}, we adopt 300-epoch training. The AdamW optimizer~\cite{kingma2014adam} is used. The learning rate, weight decay, and batch size are set default as 1.2e-3, 0.05, and 2,048, respectively. For drop path rate, we select the best of $\{0.1,0.2,0.3,0.4\}$.

We consider 5 pre-training methods of DINO~\cite{caron2021emerging}, EsViT~\cite{li2021esvit}, CLIP~\cite{radford2021clip}, DeiT~\cite{touvron2020deit}, and MAE~\cite{MaskedAutoencoders2021}, using their public checkpoints. For DINO, EsViT, and DeiT, we use their largest available models, ViT-B or Swin-B. For CLIP method, we tried both ViT-B and ViT-L.

\paragraph{Evaluation settings}

We consider 3 evaluation benchmarks: ImageNet-1K fine-tuning, ImageNet-1K linear probe, and ADE20K semantic segmentation.

\noindent \emph{ImageNet-1K fine-tuning}. We follow~\cite{bao2021beit} to use the AdamW optimizer~\cite{kingma2014adam} with layer decayed learning rates. For ViT-B, we fine-tune it by 100 epochs, and our default hyper-parameters are: batch size 2,048, learning rate 5e-3, weight decay 0.05, layer decay 0.65, and drop path rate 0.3. For ViT-L, we fine-tune it by 50 epochs, and our default hyper-parameters are: batch size 2,048, learning rate 1e-3, layer decay 0.75, and drop path rate 0.4.

\noindent \emph{ImageNet-1K linear probe}. We follow~\cite{MaskedAutoencoders2021} to use the LARS optimizer~\cite{you2017large} with a base learning rate of 0.1 and a weight decay of 0. For ViT-B, we train for 90 epochs. For ViT-L, we train for 50 epochs.

\noindent \emph{ADE20K semantic segmentation}. We follow~\cite{liu2021swin} to use an UPerNet framework~\cite{xiao2018upernet} for experiments. The AdamW \cite{kingma2014adam} optimizer is employed with the training length of 160K, a batch size of 16, and a weight decay of 0.05. For ViT-B, other hyper-parameters are set as: learning rate 4e-4, layer decay 0.65, and drop path rate 0.2. For ViT-L, other hyper-parameters are set as: learning rate 1e-4, layer decay 0.75, and drop path rate 0.4. In training, the input image size is $512\times 512$. In inference, we follow the single-scale testing of 
\cite{liu2021swin}.

\subsection{Main Results}

Table~\ref{tab:teaser_FD} shows the main results. With the feature distillation method, all listed pre-trained models are improved by 1.0\%$\sim$2.0\% on ImageNet-1K fine-tuning and 1.0$\sim$3.3 mIoU on ADE20K semantic segmentation. Particularly, by this approach, we improve the CLIP ViT-L model to reach 89.0\% on ImageNet-1K classification, which is +1.9\% higher than a sophisticated fine-tuning approach dedicated for CLIP models~\cite{ftclip2021}. 

We also improve the 3-billion-parameter SwinV2-G to achieve \textbf{61.4 mIoU} and \textbf{64.2 mAP} on ADE20K semantic segmentation and COCO object detection (using the same UperNet / HTC++ framework and the same evaluation settings as the original Swin V2 paper~\cite{swinv2}), creating new records that is +0.6 mIoU and +0.9 mAP higher than previous state-of-the-art reported in (Mask) DINO~\cite{zhang2022dino,li2022mask}, respectively, as shown in Table~\ref{tab:FD_swinv2_G}.

These results suggest the general applicability of our approach.

\subsection{Ablations}

In this section, we ablate the designs described in Section~\ref{sec:FD}. All experiments were performed on ImageNet-1K training images using ViT-B and 100-epoch training.

\paragraph{On distilling targets} Table~\ref{tab:ablation_targets} ablates the effects of different distilling targets. The use of full feature map performs best for all pre-training methods. The distillation of the full feature map can maintain more information involved in the teacher model than using other reduced features, which may lead to its superior performance. We also experimented with the classical logit distillation method on the DeiT pretrained model. It turns out that logit distillation has only +0.1\% improvements compared to the original representation. In addition, for pre-training models that do not rely on classification, the logit distillation method will not be applicable.

\paragraph{The effect of normalizing teacher features} Table~\ref{tab:ablation_normalization} ablates the effect of whether and how to perform teacher feature map normalization. Whitening the teacher feature map brings +0.8\%, +0.2\%, and +1.0\% improvements over using the original feature maps, for CLIP, DINO, and DeiT, respectively. Comparing two normalization approaches of $\ell_2$ and whitening, the whitening approach performs notably better (+0.4\%, +0.0\%, and +1.1\% for CLIP, DINO, and DeiT, respectively). Using feature map normalization also makes hyper-parameters insensitive to the pre-training models.

\begin{figure}
\centering    
    \begin{minipage}{.18\textwidth}
          \centering
        \begin{subfigure}{1.0\textwidth}
          \centering
          \includegraphics[width=0.95\linewidth]{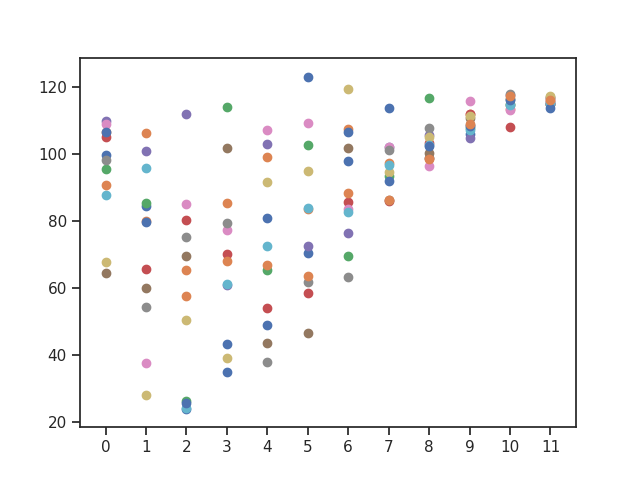}
          \caption{APE}
          \label{fig:att_patterns_a}
        \end{subfigure}%
        
        \begin{subfigure}{1.0\textwidth}
          \centering
          \includegraphics[width=0.95\linewidth]{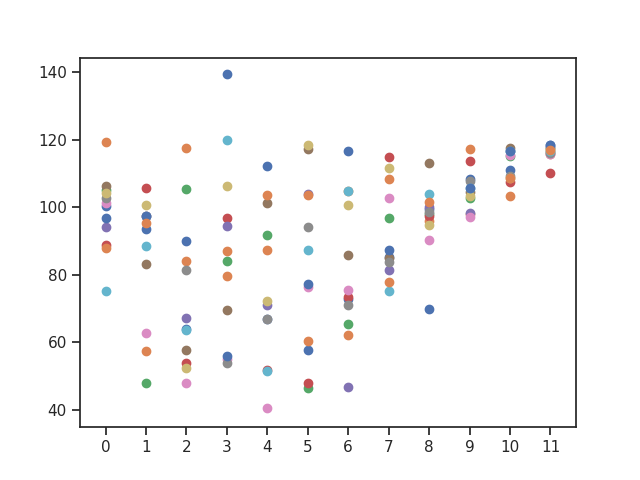}
          \caption{Non shared RPB}
          \label{fig:att_patterns_b}
        \end{subfigure}
    \end{minipage}  
    \begin{minipage}{.80\textwidth}
          \centering
        \begin{subfigure}{.5\textwidth}
          \centering
          \includegraphics[width=0.95\linewidth]{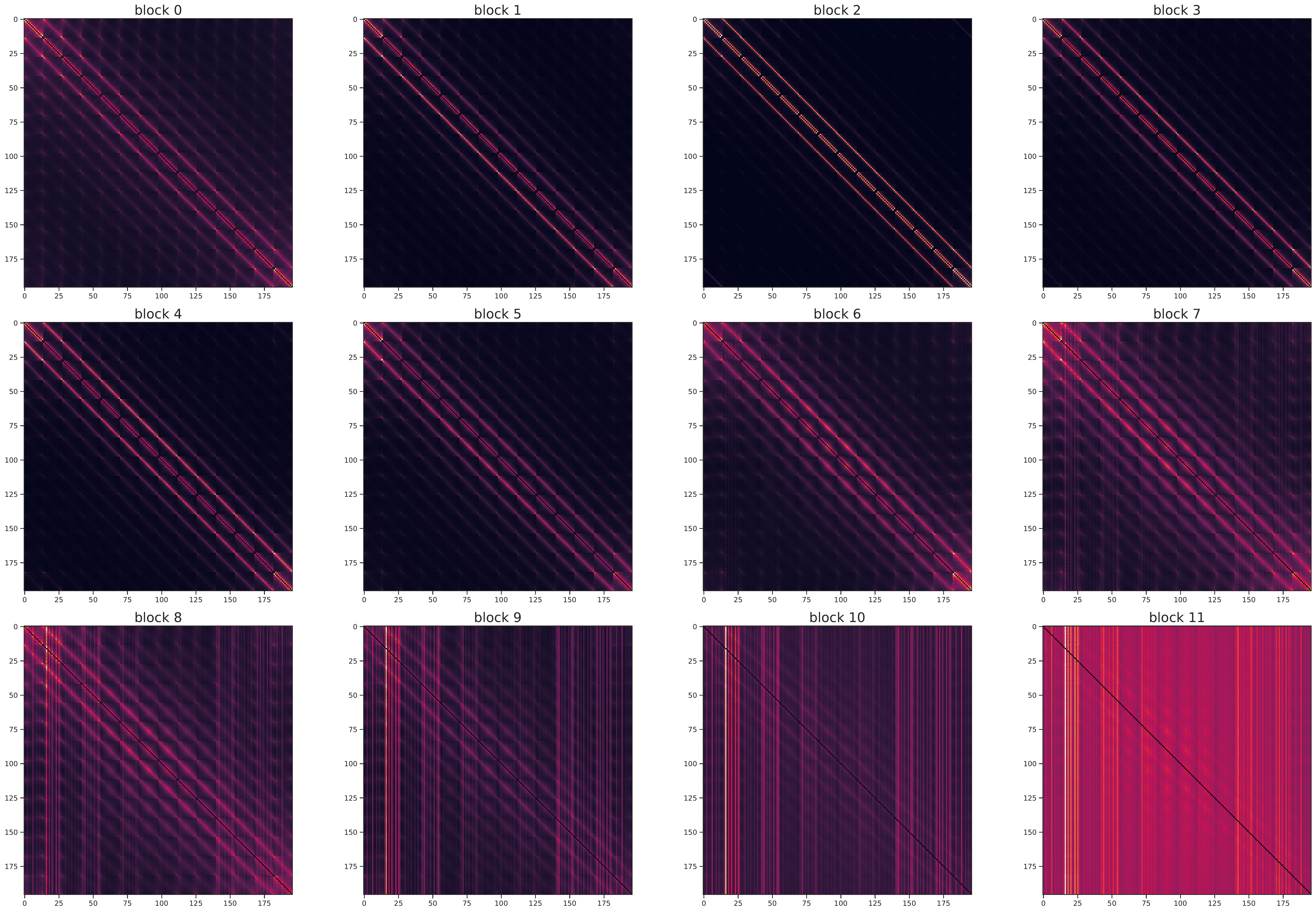}
          \caption{APE}
          \label{fig:att_patterns_a}
        \end{subfigure}%
        \begin{subfigure}{.5\textwidth}
          \centering
          \includegraphics[width=0.95\linewidth]{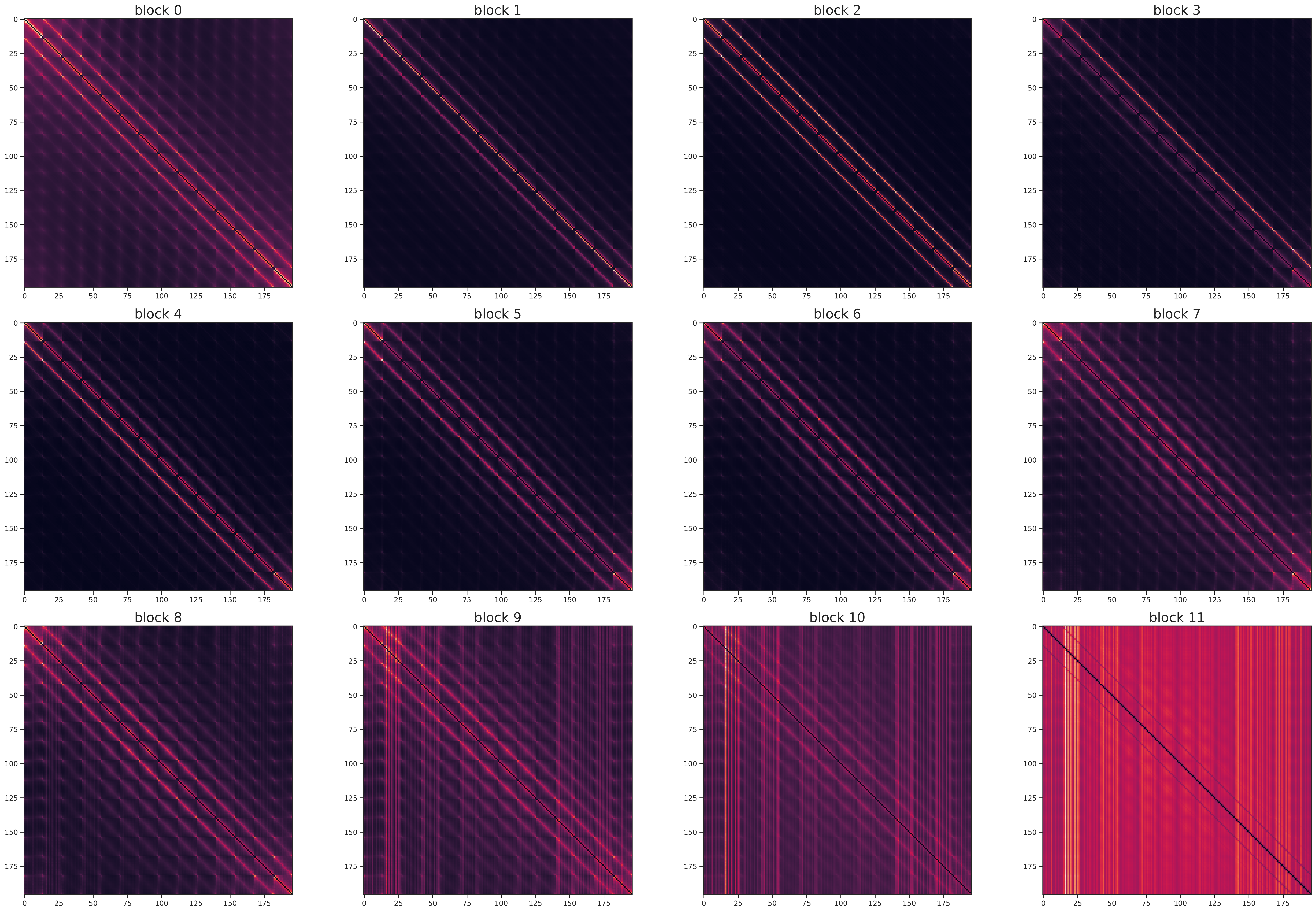}
          \caption{Non shared RPB}
          \label{fig:att_patterns_b}
        \end{subfigure}
    \end{minipage}  
    \caption{The average attention distance and average attention maps with different position encoding configurations on the student network. }
    \label{fig:position_configuration}
\end{figure}

\paragraph{On position encoding configurations} Table~\ref{tab:ablation_position} ablates the effect of varying position encoding configurations in the student network. The shared relative position bias (RPB) configuration performs best overall. Figure~\ref{fig:att_patterns} and \ref{fig:position_configuration} shows that the shared RPB configuration has the most diverse attention heads, which may result in the best accuracy. Also note that all of these configurations perform quite well, suggesting that the primary factor for success comes not from the proper position encoding configuration, but from the feature distillation algorithm itself.

\paragraph{On asymmetric drop path rates} Table~\ref{tab:ablation_dpr} ablates the effect of different degrees of drop path regularization. Moderately increasing the drop path rate of the student network would be beneficial, possibly due to the relief of over-fitting. The optimal student drop path rate for CLIP, DINO, and DeiT are 0.1, 0.2, and 0.2, respectively. For teacher networks, adding drop path regularization often compromises the performance, indicating that an accurate teacher signal is beneficial. Therefore, we adopt this asymmetric drop path rate strategy by default.

\begin{table}[t]
    \begin{minipage}{0.49\linewidth}
    \centering
    \addtolength{\tabcolsep}{-1pt}
    \caption{Ablation on distilling targets.}
    \begin{tabular}{c|ccc}
        \Xhline{2\arrayrulewidth}
            targets & CLIP & DINO & DeiT \\
            \hline
            before distill & 82.9 & 82.8 & 81.8 \\
            \hline
            logit & - & - & 81.9 \\
            CLS token & 81.7 & 82.2 &  78.0 \\
            GAP feature & 83.4 & 82.8 &  82.0 \\
            Full Feature map & 84.3 & 83.6 & 82.7 \\
            \Xhline{2\arrayrulewidth}
    \end{tabular}
    \label{tab:ablation_targets}
    
    \vspace{1em}
    
    \addtolength{\tabcolsep}{-1pt}
    \caption{The effect of normalizing teacher features.}
    \begin{tabular}{c|ccc}
        \Xhline{2\arrayrulewidth}
            normalize method & CLIP & DINO & DeiT \\
            \hline
            none & 83.5 & 83.4 & 81.7  \\
            $\ell_2$ norm & 83.9 & 83.6 & 81.6  \\
            whiten & 84.3 & 83.6 & 82.7 \\
            \Xhline{2\arrayrulewidth}
    \end{tabular}
    \label{tab:ablation_normalization}
        
\end{minipage}\hfill
\begin{minipage}{0.49\linewidth}
    \centering

    \addtolength{\tabcolsep}{-1pt}
    \caption{Ablation on position encoding configurations.}
    \begin{tabular}{c|ccc}
        \Xhline{2\arrayrulewidth}
            Position config. & CLIP & DINO & DeiT \\
            \hline
            before distill & 82.9 & 82.8 & 81.8 \\
            \hline
            APE & 84.0 & 83.4 & 82.3  \\
            non-shared RPB & 83.9 & 83.6 & 82.6 \\
            shared RPB & 84.3 & 83.6 & 82.7 \\
            \Xhline{2\arrayrulewidth}
    \end{tabular}
    \label{tab:ablation_position}
    
    \vspace{1em}
    \addtolength{\tabcolsep}{-2pt}
    \caption{Ablation on asymmetric drop path rate.}
    \begin{tabular}{cc|ccc}
        \Xhline{2\arrayrulewidth}
            stu. d.p.r. & tea. d.p.r. & CLIP & DINO & DeiT \\
            \hline
            0.1 & 0 & 84.3 & 83.2 & 82.4 \\
            0.2 & 0 & 84.0 & 83.6 & 82.7 \\
            0.3 & 0 & - & 83.6 & 81.9 \\
            best & 0.1 & 84.0 & 83.6 & - \\
            \Xhline{2\arrayrulewidth}
    \end{tabular}
    \label{tab:ablation_dpr}
    
\end{minipage}
\end{table}

\section{Discussion in the Context of Related Works}

Representation learning is a major theme in computer vision. In this section, we will discuss its goals, evaluations, property requirements, existing methods, and some reflections based on the findings of this paper.

\paragraph{Goals and evaluations of representation learning} The aims and common evaluations for representation learning include:
\begin{itemize}
\item \emph{Fine-tuning purpose}. The pioneer work~\cite{HinSal06} discovered that the layer-by-layer greedily learnt model weights can play good initialization for the final joint learning with all layers, which has actually triggered the renaissance of deep learning in 2006. Since 2012, the use of representation learning for fine-tuning has become more de facto standard, where commonly initialized using model weights trained from the ImageNet-1K image classification task has been a key factor in generalizing deep learning into a wide range of visual tasks~\cite{rcnn13,long2015fully}. This usage is so important that solid improvements in this direction can usually attract significant attention, which is right the case for the recent masked image modeling methods~\cite{bao2021beit, MaskedAutoencoders2021, xie2021simmim}.

\item \emph{Linear evaluation}. Early self-supervised learning methods~\cite{dosovitskiy2014exemplarcnn, pathak2016context, he2019moco} are usually evaluated using this protocol, which fixes the visual backbone and learns a linear classifier on top of the fixed backbone. The protocol mainly reflects the linear separability of learnt features. Since it is not directly targeted at real-world scenarios, it is not as popular as it used to be. 

\item \emph{Few-shot scenarios}. 
Human beings are very good at few-shot learning. Few-shot learning also finds numerous applications to help us to quickly build new capabilities with limited data. Self-supervised approaches such as instance contrastive learning and visual-text contrastive learning have been proven to be good for few-shot learning~\cite{wu2018memorybank,chen2020big,radford2021clip}.

\item \emph{Zero-shot / prompt-aided inference}. Another use of representation learning is zero-shot or prompt-aided inference. This allows to use pre-trained models without additional training on down-stream data~\cite{radford2021clip,gu2021open,xu2021simple}. This use of representation learning makes possible for a single model to cope with all tasks, and it is now attracting more and more attention from the community. 

\end{itemize}

This paper is mainly concerned with the fine-tuning usage of representation learning. We will discuss:

\paragraph{What are nice properties for fine-tuning?} Few studies have attempted to study which properties are good for fine-tuning. In this discussion, we'll divide the potentially nice properties as: 

\begin{itemize}
    \item \emph{Optimization friendliness}. The pre-trained model is used as the initialization of the fine-tuning tasks. Good initialization can help the optimization process in fine-tuning. Given the same optimizer in fine-tuning, a better initialization may result in flatter loss landscape, and better accuracy~\cite{losslandscape2017}. 
    \item \emph{Encoding of scalable and generalizable knowledge}. Pre-trained models are supposed to encode knowledge that can well accomplish the pre-training task. The knowledge encoded in the pre-trained models may be generally beneficial for down-stream tasks, for example, the semantics encoded in the CLIP models turn out to be significantly helpful for down-stream tasks such as ImageNet-1K classification and ADE20K segmentation tasks. Another nice property is the scalability to encode the knowledge: Can models encode well the richer knowledge when we have larger data? Can larger models be well driven by the learning task? This is often discussed in the context of \emph{scaling law}~\cite{kaplan2020scaling}, which has become a corner belief behind the recent remarkable success of natural language models. We expect this also applies to computer vision.
\end{itemize}

The feature distillation approach introduced in this paper mainly improves optimization friendliness as analyzed throughout this paper. 
We hope it to provide a way for the study of representation learning to focus more effort on the second nice property of generality and scalability. In the following, we will make some reflections on existing representation learning approaches in the context that there has been a general method such as feature distillation to improve the optimization friendliness.

\paragraph{Existing representation learning approaches and our reflections}

There are four notable representation learning approaches, including image classification~\cite{alexnet} on ImageNet-1K~\cite{deng2009imagenet}, instance contrastive learning~\cite{dosovitskiy2014exemplarcnn,he2019moco}, visual-text contrastive learning~\cite{radford2021clip,jia2021align}, and masked image modeling~\cite{bao2021beit,xie2021simmim,MaskedAutoencoders2021}. We made the following reflections to these approaches:
\begin{itemize}
   \item \emph{Masked image modeling (MIM)}. Masked image modeling has attracted a lot of attention for its excellence in fine-tuning evaluations. As this paper suggests, the excellent performance may come primarily from the optimization friendliness of the learnt representation. 
   In terms of its scalability, the ability of MIM tasks to train large-capacity models has been well demonstrated in~\cite{MaskedAutoencoders2021,xie2021simmim,swinv2}, however, the ability to benefit from larger data currently sounds negative, as shown in~\cite{el2021large}. We think that this issue, if not well addressed, could hinder its further popularity.
   \item \emph{Instance contrastive learning}. The instance contrastive learning method performs pre-training in a self-supervised manner, which has attracted a lot of attention since it surpassed the supervised classification method on multiple down-stream tasks~\cite{he2019moco}. Moreover, it achieved impressive accuracy using linear and few-shot evaluations~\cite{chen2020big}. However, when it comes to vision Transformer backbones~\cite{dosovitskiy2020vit,liu2021swin}, the fine-tuning performance becomes inferior to others~\cite{xie2021self,li2021esvit}, which may have hindered its appeal for now. Another issue is that its scalability to model capacity and data size is rarely studied, or is only performed in a linear evaluation setup~\cite{Tian_2021_ICCV}. Therefore, its actual behavior on the scalability property is unclear. We hope that our significant improvements to its fine-tuning performance will encourage the community to reinvigorate the research on this method. It will be interesting to see a solid study about its scalability to model capacity and data size.
   \item \emph{Image classification and visual-text contrastive learning}. 
   Image classification has been the standard upstream pre-training task for nearly a decade since AlexNet~\cite{alexnet}. The visual-text alignment task opens up the field of zero-shot recognition. We now discuss them within a single item because they can basically be formulated with the same objective~\cite{wei2022icar,yang2022unified}, and they possess some similar properties. We first examine existing studies about its scalability to model capacity and data size. In fact, both of these tasks have been shown to perform scalable~\cite{zhai2021scaling, swinv2, radford2021clip, jia2021align, radford2021clip, jia2021align}. In general, the data for the visual-text tasks is more readily available and thus is more friendly for the data-hungry large model training. Moreover, the fine-tuning performance is also shown able to be significantly enhanced with the aid of our feature distillation method, which would further make this pre-training task more attractive. In these considerations, this task would be a good choice for large-scale representation learning. It will be also interesting to see if it can be combined with other representation learning methods to improve its data mining efficiency. 
\end{itemize}

There are many other representation learning approaches such as: gray-scale image colorization~\cite{zhang2016colorization}, jigsaw puzzle solving~\cite{noroozi2016jigsaw}, split-brain auto-encoding~\cite{zhang2017splitbrain}, rotation prediction~\cite{gidaris2018rotation}, learning to cluster~ \cite{caron2018deepcluster}, or predicting values of some channels with one or two other channels as input~\cite{zhang2016colorization,zhang2017splitbrain}, and pixel-level contrast~\cite{xie2021pixpro,wang2020DenseCL}. In the context of our feature distillation approach, some of them may worth re-examined.

\section{Conclusion}

This paper has introduced a simple feature distillation approach that can generally improve the fine-tuning performance of many visual pre-training models. It made the contrastive based self-supervised learning methods as competitive in fine-tuning as the state-of-the-art masked image modeling (MIM). It also improved a CLIP pre-trained ViT-L model to reach 89.0\% top-1 accuracy on ImageNet-1K classification. While our analysis through a set of attention- and optimization-related diagnosis tools suggests that our feature distillation approach mainly improves the optimization friendliness of learnt representations, we hope the findings can provide a way for the future research to focus more effort on the generality and scalability of the learnt representations.

\section*{Acknowledgement}

All SwinV2-G experiments are conducted by Ze Liu.

%\newpage
\bibliographystyle{apalike}
\bibliography{ref}

\newpage

\appendix
\section{Average Attention Maps for DINO and DeiT}

\begin{figure}[h]
\centering
\begin{subfigure}{.48\textwidth}
  \centering
  \includegraphics[width=0.95\linewidth]{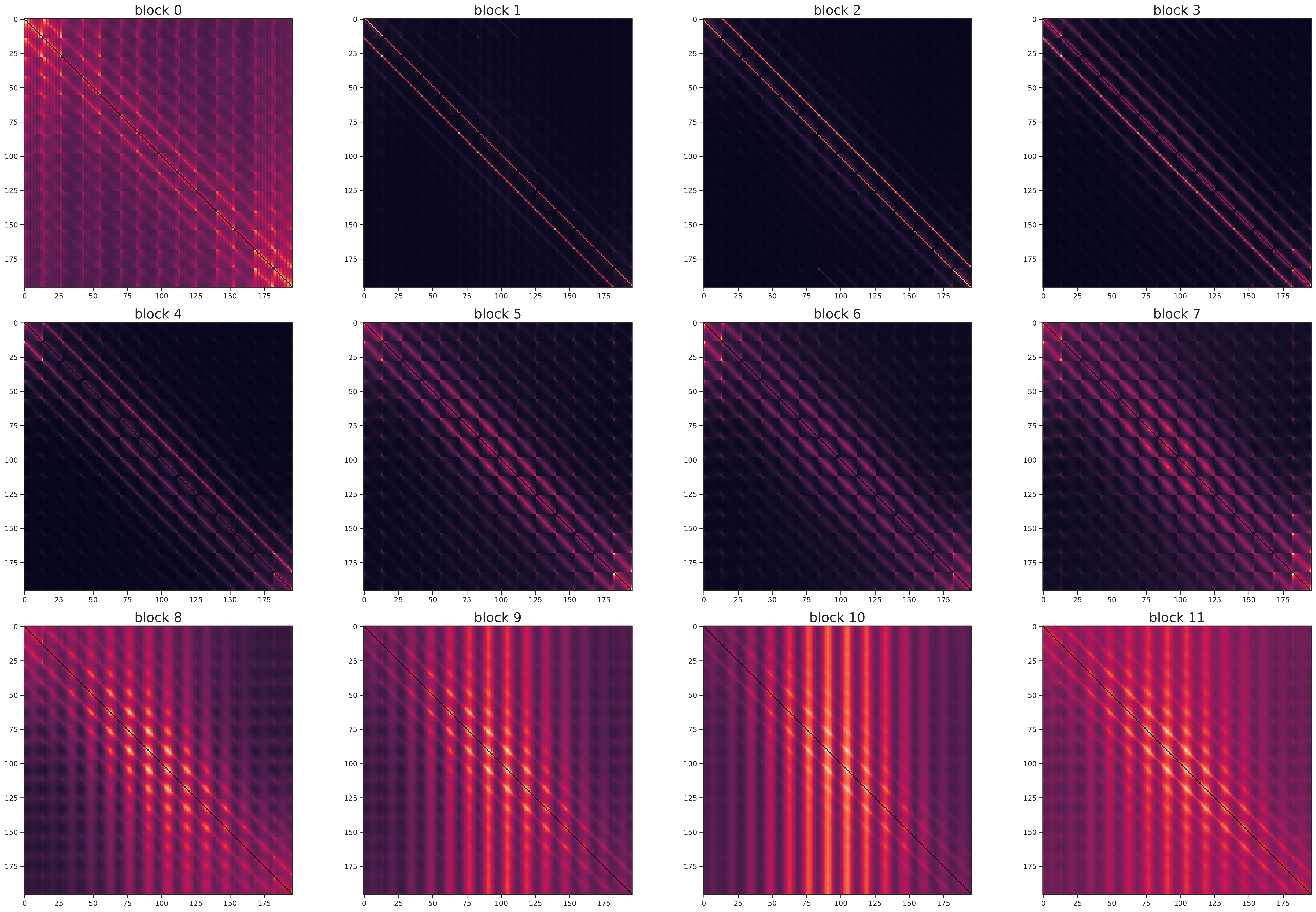}
  \caption{Before feature distillation}
  \label{fig:att_patterns_a}
\end{subfigure}
\begin{subfigure}{.48\textwidth}
  \centering
  \includegraphics[width=0.95\linewidth]{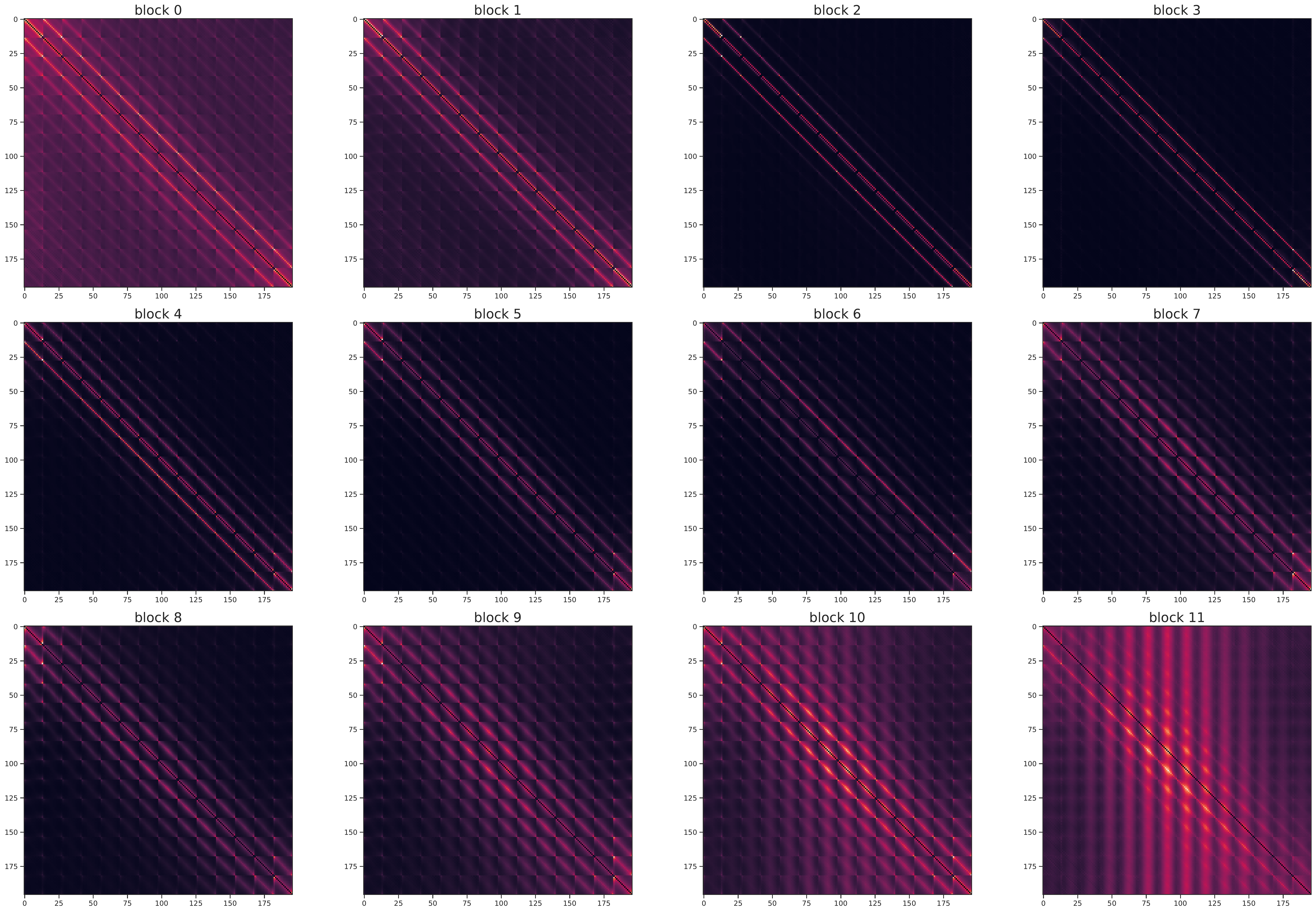}
  \caption{After feature distillation}
  \label{fig:att_patterns_b}
\end{subfigure}
    \caption{Average attention maps per layer using the DINO ViT-B model before and after feature distillation. The image patches are indexed starting from top-left to bottom-right. The 12 layers' average attention maps (Layer 0-11) are visualized from top-left to bottom-right.}
    \label{fig:att_patterns_dino}
\end{figure}

\begin{figure}[h]
\centering
\begin{subfigure}{.48\textwidth}
  \centering
  \includegraphics[width=0.95\linewidth]{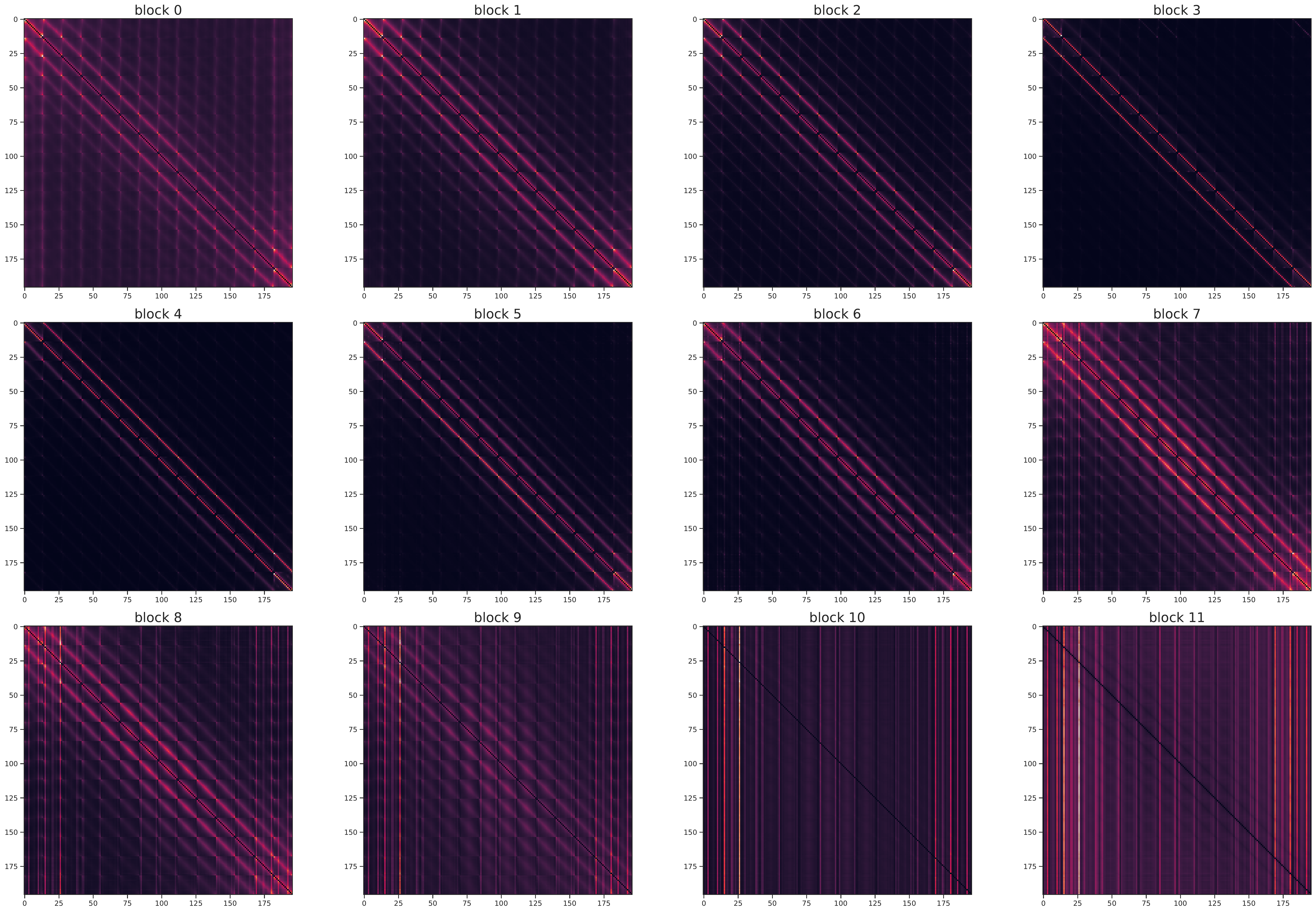}
  \caption{Before feature distillation}
  \label{fig:att_patterns_a}
\end{subfigure}
\begin{subfigure}{.48\textwidth}
  \centering
  \includegraphics[width=0.95\linewidth]{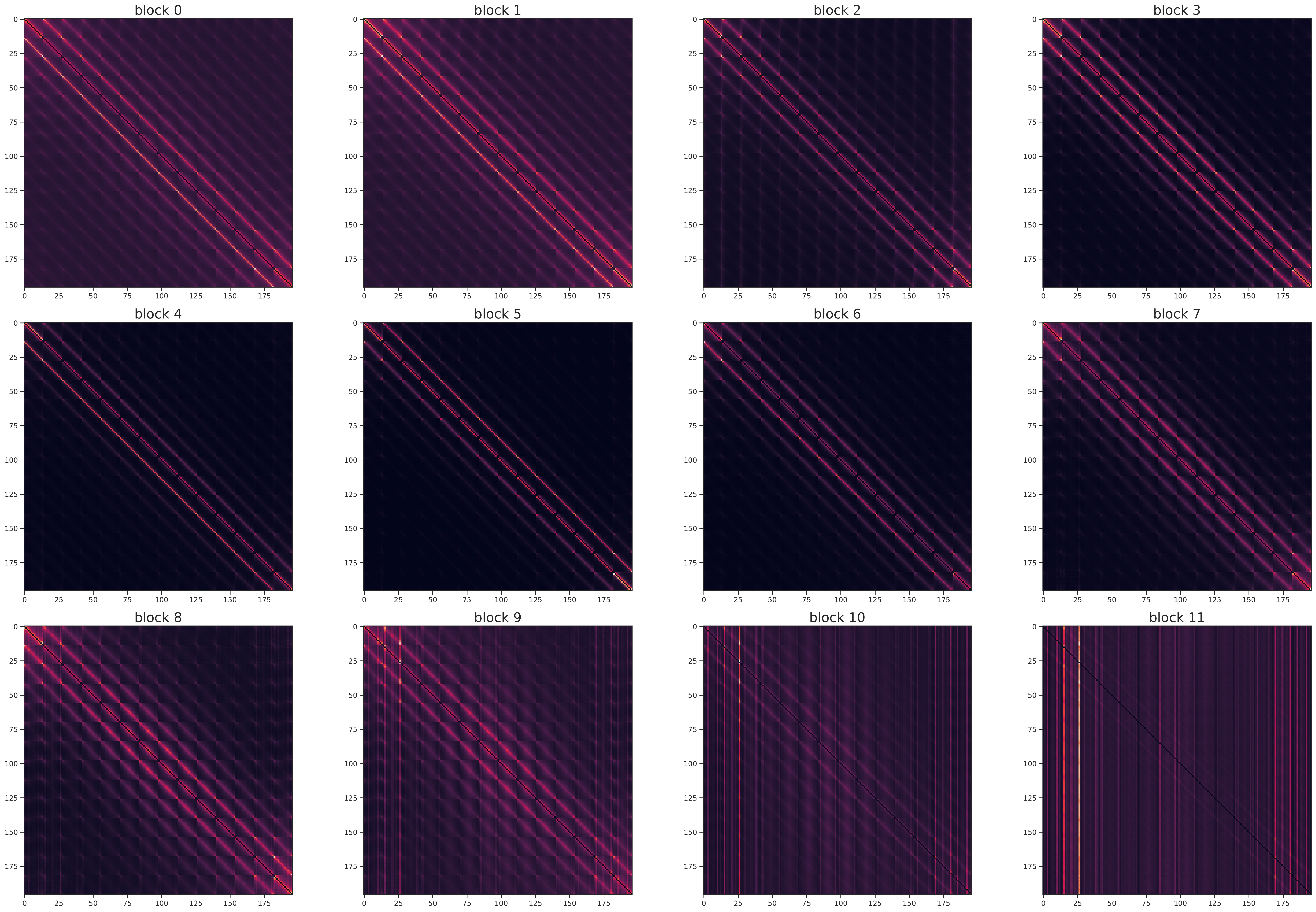}
  \caption{After feature distillation}
  \label{fig:att_patterns_b}
\end{subfigure}
    \caption{Average attention maps per layer using the DeiT ViT-B model before and after feature distillation. The image patches are indexed starting from top-left to bottom-right. The 12 layers' average attention maps (Layer 0-11) are visualized from top-left to bottom-right.}
    \label{fig:att_patterns_deit}
\end{figure}

In the main paper, we present the average attention maps of the CLIP pre-trained ViT-B model before and after feature distillation. In Figure~\ref{fig:att_patterns_dino} and~\ref{fig:att_patterns_deit}, we illustrate the maps for DINO and DeiT pre-trained ViT-B models, respectively. Similar to the observations for the CLIP pre-trained model, DINO and DeiT models after feature distillation are also observed with more \emph{diagonal} patterns than before. This indicates that our feature distillation approach can generally enhance the translational invariance properties for various pre-training models.

\section{Hyperparameters for Feature Distillation}

\begin{table}[H]
    \centering
    \caption{
    Hyperparameters for feature distillation on ImageNet-1K.
    }
    \begin{tabular}{l|cc}
    \toprule
        \bf Hyperparameters & \bf Base Size & \bf Large Size \\
        \midrule
            Patch size & $16 \times 16$ & $14 \times 14$ \\
            Layers  & 12 & 24 \\
            Hidden size & 768 & 1024 \\
            FFN inner hidden size & 3072 & 4096 \\
            Attention heads & 12 & 16 \\
            Attention head size & \multicolumn{2}{c}{64} \\
        \midrule
            Training epochs & \multicolumn{2}{c}{300} \\
            Batch size & \multicolumn{2}{c}{2048} \\
            Adam $\epsilon$ & \multicolumn{2}{c}{1e-8} \\
            Adam $\beta$ & \multicolumn{2}{c}{(0.9, 0.999)} \\
            Peak learning rate & \multicolumn{2}{c}{1.2e-3} \\
            Minimal learning rate & \multicolumn{2}{c}{2e-5} \\
            Learning rate schedule & \multicolumn{2}{c}{Cosine} \\
            Warmup epochs & \multicolumn{2}{c}{10} \\
        \midrule
            Gradient clipping & \multicolumn{2}{c}{3.0} \\
            Dropout & \multicolumn{2}{c}{\xmark} \\
            Weight decay & \multicolumn{2}{c}{0.05} \\
            Stoch. depth & \{0.1,0.2,0.3\} & 0.3 \\
        \midrule
            Data Augment & \multicolumn{2}{c}{RandomResizeAndCrop 0.08-1} \\
            Input resolution & \multicolumn{2}{c}{$224 \times 224$} \\
        \bottomrule
    \end{tabular}
    \label{tab:appd-hyper-pretrain}
\end{table}

\section{Hyperparameters for Fine-tuning}

\begin{table}[H]
    \centering
    \caption{
    Hyperparameters for fine-tuning on ImageNet-1K.
    }
    \begin{tabular}{l|cc}
    \toprule
    \bf Hyperparameters & \bf Base Size & \bf Large Size \\
    \toprule
    Peak learning rate & {\{5e-3, 6e-3\}} & 1e-3 \\ 
    Fine-tuning epochs & 100 & 50 \\
    Warmup epochs  & 20 & 5 \\
    Layer-wise learning rate decay & {\{0.6, 0.65\}} & 0.75 \\
    Batch size & \multicolumn{2}{c}{2048} \\
    Adam $\epsilon$ & \multicolumn{2}{c}{1e-8}  \\
    Adam $\beta$ & \multicolumn{2}{c}{(0.9, 0.999)} \\
    Minimal learning rate & \multicolumn{2}{c}{2e-6} \\
    Learning rate schedule & \multicolumn{2}{c}{Cosine} \\
    \midrule
    Repeated Aug & \multicolumn{2}{c}{\xmark} \\
    Weight decay & \multicolumn{2}{c}{0.05} \\
    Label smoothing $\varepsilon$ & \multicolumn{2}{c}{0.1}     \\
    Stoch. depth & \{0.1,0.2,0.3\} & 0.4 \\
    Dropout & \multicolumn{2}{c}{\xmark} \\
    Gradient clipping & \multicolumn{2}{c}{5.0} \\
    \midrule
    Erasing prob.  & \multicolumn{2}{c}{0.25} \\
    Input resolution & \multicolumn{2}{c}{$224 \times 224$} \\
    Rand Augment  & \multicolumn{2}{c}{9/0.5} \\
    Mixup prob.  & \multicolumn{2}{c}{0.8}     \\
    Cutmix prob.   & \multicolumn{2}{c}{1.0}    \\
    Color jitter & \multicolumn{2}{c}{0.4} \\
    \bottomrule
    \end{tabular}
    \label{tab:appd-hyper-finetune}
\end{table}

\section{Representation Properties of Masked Image Modeling before and after Distillation}
\begin{figure}[h]
\centering
\begin{subfigure}{.47\textwidth}
  \centering
  \includegraphics[width=0.95\linewidth]{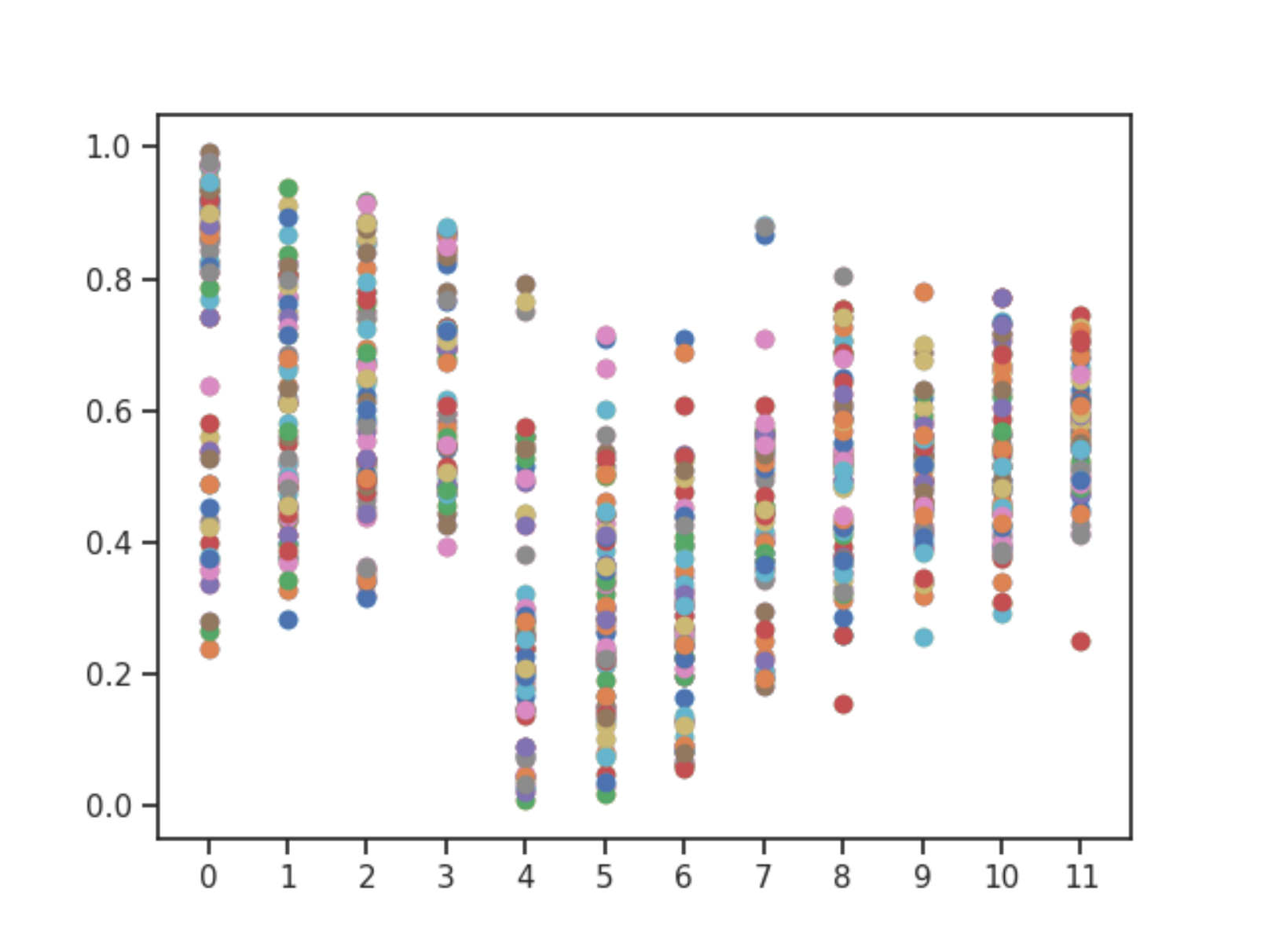}
  \caption{Before feature distillation}
  \label{fig:att_patterns_a}
\end{subfigure}%
\begin{subfigure}{.47\textwidth}
  \centering
  \includegraphics[width=0.95\linewidth]{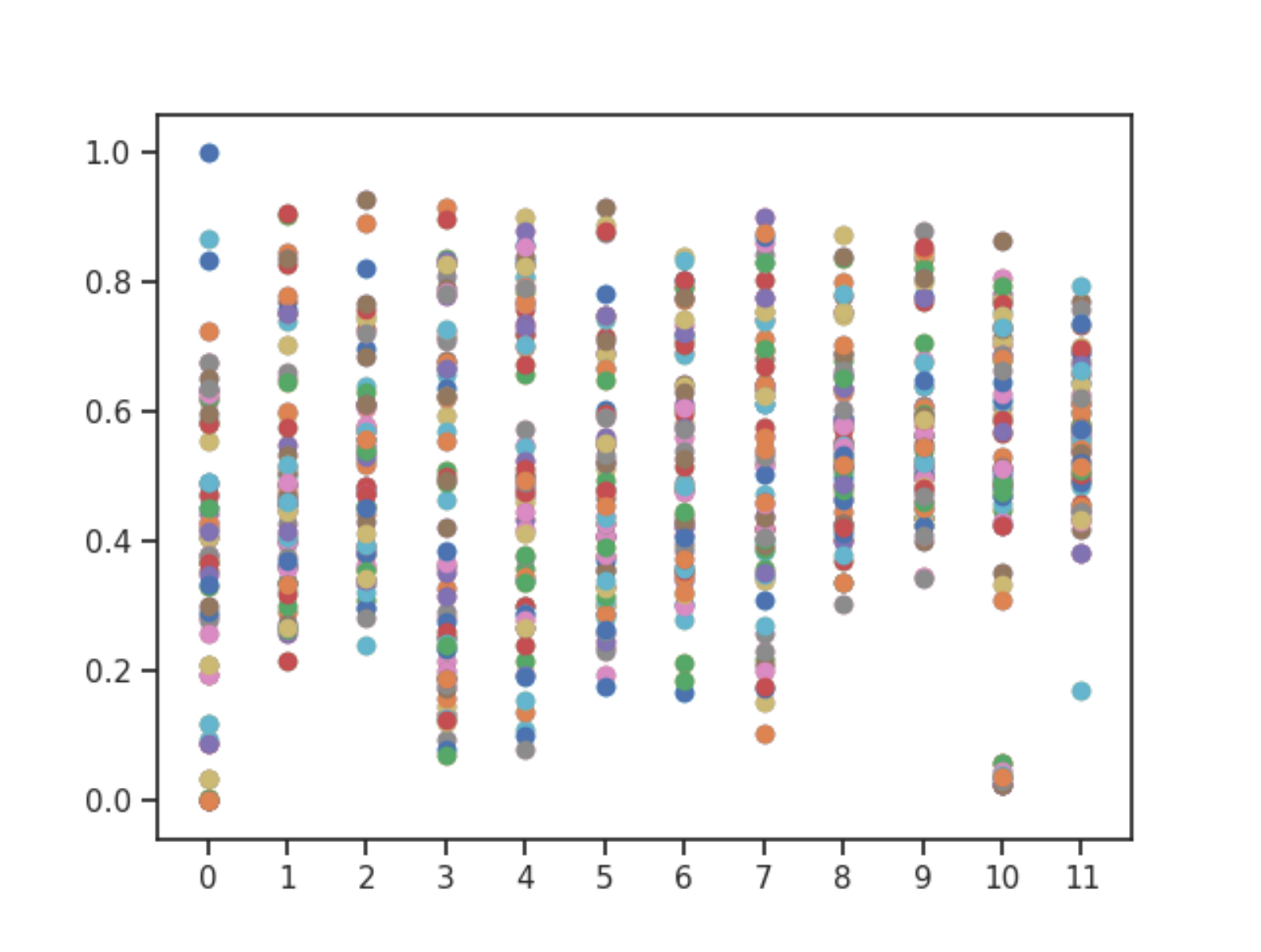}
  \caption{After feature distillation}
  \label{fig:att_patterns_b}
\end{subfigure}
    \caption{Average cosine similarity of attention maps between heads per layer before (left) and after (right) feature distillation using the MAE ViT-B.}
    \label{fig:attn_sim_mae}
\end{figure}

In Figure~\ref{fig:fd_mae}, we show the average attention distance and the loss accuracy landscapes of MAE before and after the feature distillation process. We further include the other two diagnosis tools, average attention maps in Figure~\ref{fig:attn_patterns_mae_appendix} and average attention head cosine similarity in Figure~\ref{fig:attn_sim_mae}. These two diagnosis tools also show similar behaviors before and after the feature distillation process, which also suggests that the feature distillation and masked image modeling method have certain overlap in functionality.
\begin{figure}[h]
\centering
\begin{subfigure}{.47\textwidth}
  \centering
  \includegraphics[width=0.95\linewidth]{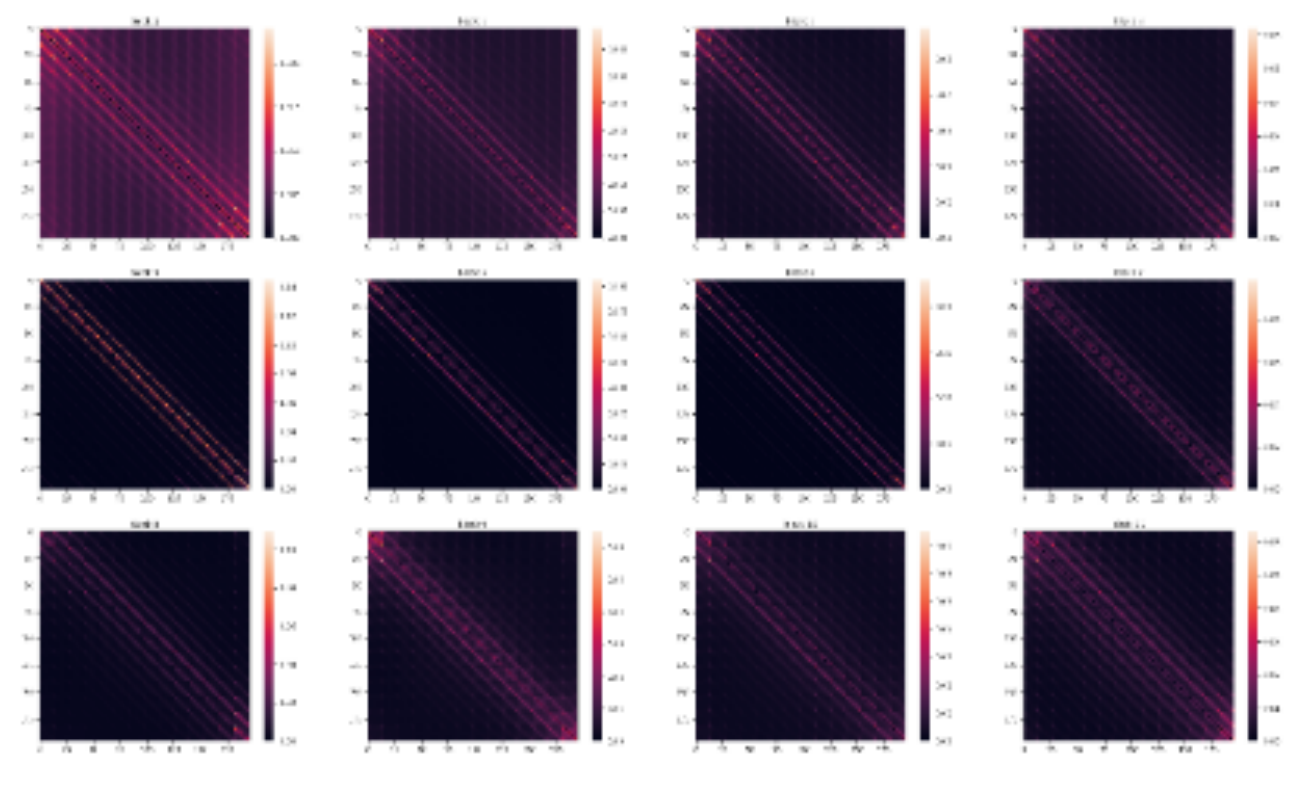}
  \caption{Before feature distillation}
  \label{fig:att_patterns_a}
\end{subfigure}
\begin{subfigure}{.47\textwidth}
  \centering
  \includegraphics[width=0.95\linewidth]{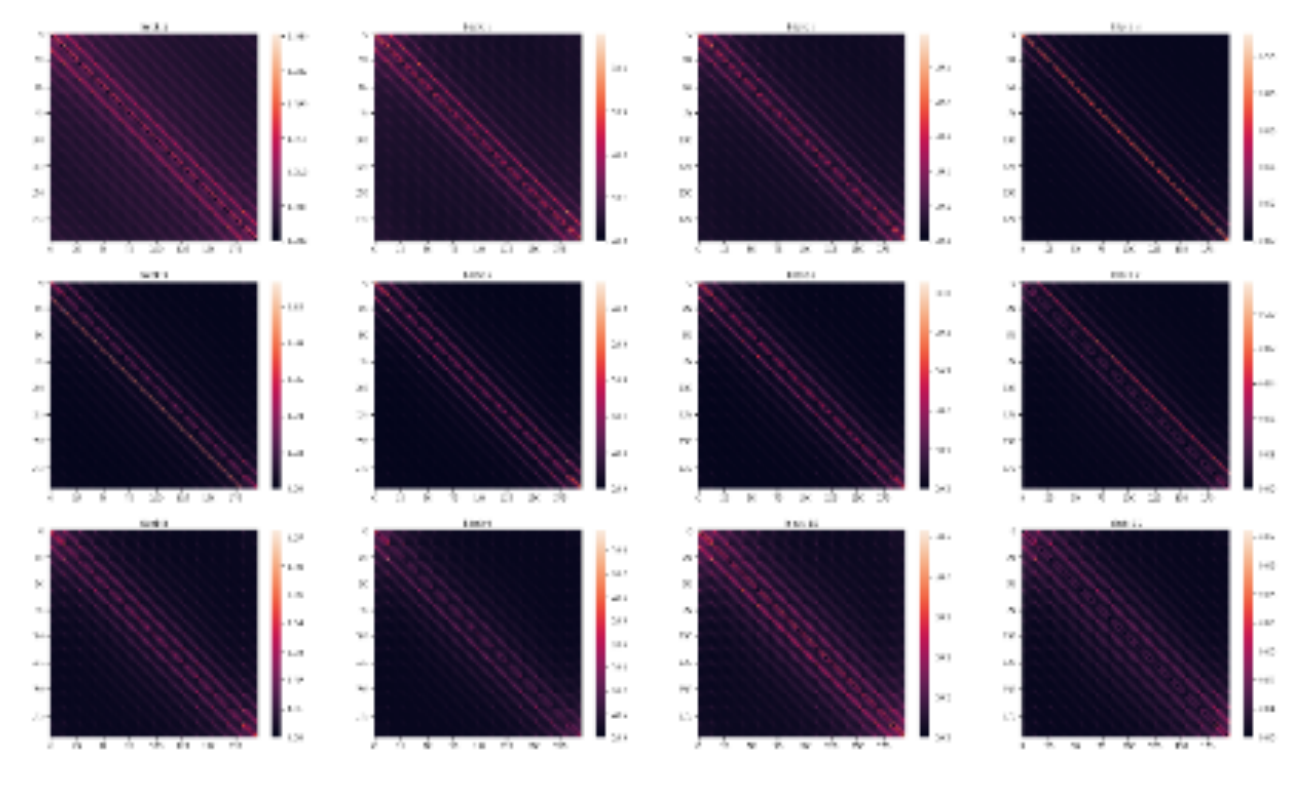}
  \caption{After feature distillation}
  \label{fig:att_patterns_b}
\end{subfigure}
    \caption{Average attention maps per layer using the MAE ViT-B model before and after feature distillation. The image patches are indexed starting from top-left to bottom-right. The 12 layers' average attention maps (Layer 0-11) are visualized from top-left to bottom-right.}
    \label{fig:attn_patterns_mae}
\end{figure}

\newpage
\section{Additional Training Cost by Feature Distillation Compared to that of Pre-training}

All experiments in Table~\ref{tab:teaser_FD} use 300-epoch ImageNet-1K training. This means that the additional training cost is constant. The additional training overhead could small if the pre-training methods are costly, e.g., about +3\% for the CLIP model. Table~\ref{tab:Efficiency} showed the equivalent training epochs of the original pre-training methods counted according to the feature distillation epochs, as well as the performance gains on ImageNet-1K image classification.

\begin{table}[h]
\caption{Comparing the equivalent pre-training epochs of different models to the feature distillation cost. The top-1 accuracy gains on ImageNet-1K image classification are also listed for reference.}
\centering
  \begin{tabular}{l|c|c|c}
  \toprule
  Method & Equivalent \#. pre-training epochs & \#. Feature distillation epochs & Accuracy gain \\
  \hline
  \multirow{2}{*}{DeiT} & \multirow{2}{*}{300} & 100 & +0.9 \\
   &  & 300 & +1.2 \\
   \hline
  \multirow{2}{*}{DINO} & \multirow{2}{*}{1,535} & 100 & +0.8 \\
    &   & 300 & +1.0 \\
   \hline
  \multirow{2}{*}{EsViT} & \multirow{2}{*}{1,535} & 100 & +0.9 \\
    &   & 300 & +1.2 \\
   \hline
  \multirow{2}{*}{CLIP} & \multirow{2}{*}{$\sim$10,000} & 100 & +1.4 \\
    &   & 300 & +2.0 \\
  \bottomrule
  \end{tabular}
\label{tab:Efficiency}
\end{table}

\end{document}